\documentclass[lettersize,journal]{IEEEtran}
\usepackage{amsmath,amsfonts}
\usepackage{algorithm}
\usepackage{array}
\usepackage[caption=false,font=normalsize,labelfont=sf,textfont=sf]{subfig}
\usepackage{textcomp}
\usepackage{stfloats}
\usepackage{url}
\usepackage{verbatim}
\usepackage{graphicx}
\usepackage{cite}
\usepackage{booktabs}
\usepackage[table]{xcolor} 

\usepackage{color, colortbl}
\definecolor{Gray}{gray}{0.9}
\newcommand\egno{\textit{e.g.}}
\newcommand\ieno{\textit{i.e.}}
\newcommand\etcno{\textit{etc}}
\usepackage{soul}
\newcommand{\modelAbbr}{COLA}
\usepackage{enumitem}
\usepackage{multirow}
\usepackage{xr}
\usepackage[normalem]{ulem}
\usepackage{algpseudocode}

\begin{document}

\title{COLA: Context-aware Language-driven \\Test-time Adaptation}

\author{Aiming Zhang$^\star$, Tianyuan Yu$^\star$, Liang Bai, Jun Tang, Yanming Guo, Yirun Ruan, Yun Zhou, 
Zhihe Lu$^\dagger$
\thanks{Aiming Zhang, Tianyuan Yu, Liang Bai, Jun Tang, Yanming Guo and Yirun Ruan are with the Laboratory for Big Data and Decision, National University of Defense Technology, Changsha, China (e-mail: aiming\_ch@nudt.edu.cn; ty.yu@nudt.edu.cn; bailiang\_nudt@163.com; tangjun06@nudt.edu.cn; guoyanming@nudt.edu.cn; ruanyirun@163.com). 
Yun Zhou is with National Key Laboratory of Information Systems Engineering, National University of Defense Technology, Changsha, China (e-mail: zhouyun@nudt.edu.cn).
Zhihe Lu is with College of Science and Engineering, Hamad Bin Khalifa University, Qatar (e-mail: zlu@hbku.edu.qa)
}
\thanks{$\star$ These authors contribute equally to this work.}
\thanks{$\dagger$ Corresponding Author}
\thanks{Manuscript received April XX, XXXX; revised August XX, XXXX.}}

\markboth{Journal of \LaTeX\ Class Files,~Vol.~14, No.~8, August~2021}%
{Shell \MakeLowercase{\textit{et al.}}: A Sample Article Using IEEEtran.cls for IEEE Journals}

\IEEEpubid{\footnotesize © 2025 IEEE. Accepted for publication in IEEE Transactions on Image Processing.}

\maketitle

\begin{abstract}
    Test-time adaptation (TTA) has gained increasing popularity due to its efficacy in addressing ``distribution shift'' issue while simultaneously protecting data privacy.
    However, most prior methods assume that a paired source domain model and target domain sharing the same label space coexist, heavily limiting their applicability.
    In this paper, we investigate a more general source model capable of adaptation to multiple target domains without needing shared labels.
    This is achieved by using a pre-trained vision-language model (VLM), \egno, CLIP, that can recognize images through matching with class descriptions.
    While the zero-shot performance of VLMs is impressive, they struggle to effectively capture the distinctive attributes of a target domain.
    To that end, we propose a novel method -- Context-aware Language-driven TTA (COLA). 
    The proposed method incorporates a lightweight context-aware module that consists of three key components: a task-aware adapter, a context-aware unit, and a residual connection unit for exploring task-specific knowledge, domain-specific knowledge from the VLM and prior knowledge of the VLM, respectively. 
    It is worth noting that the context-aware module can be seamlessly integrated into a frozen VLM, ensuring both minimal effort and parameter efficiency.
    Additionally, we introduce a Class-Balanced Pseudo-labeling (CBPL) strategy to mitigate the adverse effects caused by class imbalance.
    We demonstrate the effectiveness of our method not only in TTA scenarios but also in class generalisation tasks.
    The source code is available at \url{https://github.com/NUDT-Bai-Group/COLA-TTA}.
\end{abstract}


\bstctlcite{IEEEexample:BSTcontrol}
\section{Introduction}
\IEEEPARstart{D}{eep} neural networks have achieved remarkable performance in various areas when training and test data are drawn from the same distribution.
However, this constraint is often violated in real-world applications, \egno, person re-identification~\cite{wei2018person}, self-driving environment \cite{cordts2016cityscapes}, medical image analysis \cite{guan2021domain}, \etcno. 
The shift in distribution between training and test data can lead to a significant performance drop when directly employing a trained model on test data~\cite{patel2015visual}. 
Addressing this issue is challenging because it is impossible to accurately predict the exact distribution of the test data and collect training data accordingly.

\begin{figure}[t]
\begin{center}
\includegraphics[width=0.5\textwidth]{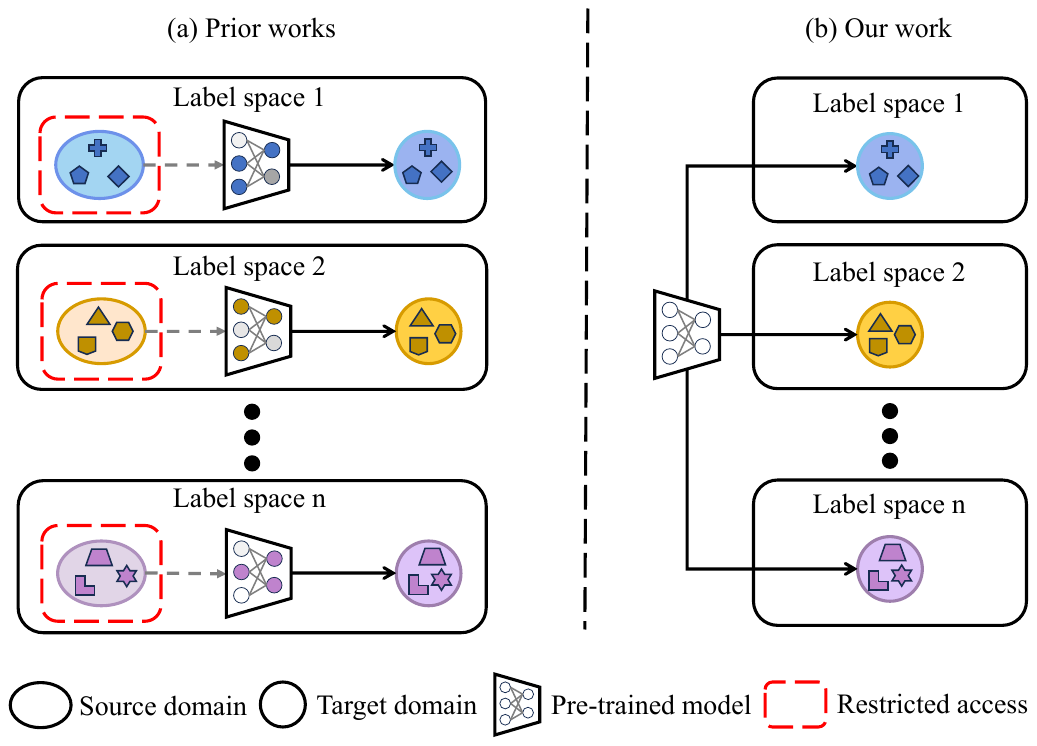}
\end{center}

\caption{Comparison between prior works and our work. (a) In prior works, source and target domains share the same label space,
leading to intensive training efforts for various models tailored for specific target domains.
(b) In contrast, our work aims at building a unified model that can be adapted to arbitrary target domains.}
\label{fig:module}

\end{figure}

To address the distribution shift issue, unsupervised domain adaptation (UDA) \cite{long2015learning,lu2020stochastic} has been proposed. UDA leverages knowledge gained from the source domain to adapt to the target domain. This adaptation process generally requires access to labeled source data and unlabeled target data simultaneously.
However, in some privacy-sensitive applications, accessing the source domain data is prohibited due to privacy concerns and business regulations. 
In such cases, test time adaptation (TTA)~\cite{liang2020we} has been proposed.
Specifically, TTA assumes that source domain data is unavailable and leverages only a model trained on the source domain for adaptation to the unlabeled target domain.
\IEEEpubidadjcol

However, most existing TTA methods \cite{Chen_2022_CVPR, gao2022_DEPT, lu2023uncertainty} are typically based on an idealized assumption, \ieno, source and target domains share the same label space, as illustrated in Figure~\ref{fig:module}(a). 
This assumption, however, hardly holds in practice because 
(i) the label space of the target domain is often uncertain and may evolve over time, making a pre-trained source model inapplicable; 
(ii) collecting a source dataset sharing the same label space as a target domain is a non-trivial task.
The emergence of large-scale vision language models (VLMs) offers a solution to break the assumption, which is achieved by recognizing an image by matching with the corresponding class description.
More importantly, these VLMs trained on million-level image-text pairs, \egno, 400M for Contrastive language-image pre-training (CLIP)~\cite{radford2021learning}, demonstrate impressive performance on an unseen dataset in a zero-shot manner, showing its great potential in TTA.

To overcome the restricted assumption of existing TTA methods and exploit the potential of VLMs, this paper proposes a novel method called Context-aware Language-driven TTA (COLA). 
COLA utilises pre-trained VLMs, such as CLIP~\cite{radford2021learning}, as a unified model adaptable to diverse target domains, as illustrated in Figure~\ref{fig:module}(b). 
Empirical observations indicate that while CLIP exhibits impressive zero-shot performance on a target domain, it falls short compared to existing methods that explore task-specific knowledge.
To adapt VLMs to the target domain with minimal effort and enhanced parameter efficiency, COLA incorporates a lightweight context-aware module (CAM) to modulate the knowledge acquired by VLMs.
CAM is specifically designed with three key components: the task-aware adapter, the context-aware unit (CAU), and the residual connection. These three parts are expected to learn various knowledge from target domain: (i) the task-aware adapter learns knowledge unique to the target domain; (ii) the CAU explores domain-specific knowledge from the VLM; (iii) the residual connection unit preserves the potential of the pre-trained VLM. 
The COLA framework harnesses the synergistic capabilities of its constituent parts to assimilate task-specific knowledge while simultaneously preserving and leveraging the rich semantic understanding encoded in VLMs.

To further enhance the optimization of COLA using high-quality pseudo-labels, we introduce a novel technique termed Class-Balanced Pseudo-Labeling (CBPL). 
Traditional pseudo-labeling approaches often employ a uniform confidence threshold to filter samples across the entire dataset. This practice 
can lead to an imbalance in the representation of samples across different classes. 
This imbalance introduces bias in the learning process, which may result in biased predictions and reduced accuracy for underrepresented classes.
In contrast, the proposed CBPL method introduces a tailored set of thresholds for each individual class, ensuring a balanced representation and thereby mitigating the risk of class imbalance.
By incorporating this pseudo-labeling strategy, our COLA further improves the ability of learning domain knowledge and effectively leverages the potential of the pre-trained VLM. 

Our contributions can be summarized as follows:
\begin{itemize}[leftmargin=*]
    \item To address the strict TTA assumption of identical label spaces between source and target domains, we introduce a unified method capable of adapting to diverse target domains even without needing a source domain.
    \item Our proposed method, Context-aware Language-driven TTA (COLA), seamlessly integrates a lightweight module -- context-aware module (CAM) with a pre-trained VLM. 
    CAM is designed to explore task-specific and domain-specific knowledge while leveraging the prior knowledge encoded in the VLMs.
    \item A CBPL method is proposed to optimize our COLA for mitigating the adverse effect of class imbalance.
    \item The proposed method achieves new state-of-the-art (SOTA) performance on five major TTA benchmarks compared with existing methods, \egno, 7.6\% improvement with the ViT-B/16 backbone on VisDA-C.
    Notably, our model is efficient in parameter count and computational cost, and remains easy to integrate into existing VLMs.
    Furthermore, the proposed method shows the superior generalization capability. 
\end{itemize}

\section{Related work}
\subsection{Test-time Adaptation}
Test-time adaptation (TTA) aims to quickly adapt a well-trained source domain model to unlabeled target data without accessing source data, as described by Liang et al.~\cite{liang2023comprehensive}. TTA encompasses several variants: test-time domain adaptation (TTDA), online test-time adaptation, test-time batch adaptation, and test-time prior adaptation. For clarity, the TTA addressed in this paper specifically refers to TTDA, also known as Source-free Domain Adaptation, and should not be confused with OTTA. Unlike OTTA, which adapts to each new test sample as it streams in, TTDA utilizes the entire unlabeled dataset from the target domain for comprehensive adaptation, addressing a broad spectrum of challenges and scenarios.

Researchers have extensively explored TTA from various perspectives, including methods based on pseudo-labeling~\cite{liang2020we,yang2021_NRC,wang2022_CDCL}, consistency~\cite{Chen_2022_CVPR}, clustering~\cite{liang2020we,wang2020tent}, source estimation~\cite{tian2021vdm,qiu2021_CPGA}, and self-supervision~\cite{liang2021shotplus,liang2023comprehensive}. Specifically, pseudo-labeling methods enhance label accuracy using semi-supervised techniques. Consistency approaches ensure stable model outputs despite variations in data or parameters. Clustering methods optimize decision boundaries by minimizing uncertain predictions or clustering target features. Source estimation aims to approximate the source domain's data distribution, transforming the problem into a well-studied domain adaptation challenge. Finally, self-supervision helps the model learn target domain features through auxiliary tasks.
Among these methods, SHOT~\cite{liang2020we} is a representative method, which integrates centroid-based pseudo-labels and maximizes mutual information for clustering. This method trains the feature extractor and classifier in the source domain, then fixes the classifier, and fine-tunes the feature extractor on the target domain. Its optimization strategy and philosophy have inspired numerous subsequent works~\cite{ahmed2022cross,lao2021hypothesis,wang2022exploring,liang2023comprehensive}.
Following this, Adacon~\cite{Chen_2022_CVPR} incorporates contrastive learning to ensure consistent predictions across different data and model variations, while enhancing model performance with advanced pseudo-label calibration and optimization techniques.
DePT~\cite{gao2022_DEPT} introduces a more advanced ViT-B/16 backbone and specifically optimizes ``visual prompts'' for the target domain, maintaining source domain knowledge during adaptation. Using a pre-trained source model and an approximate Gaussian mixture model, Tian et al.~\cite{tian2021vdm} generate virtual domain samples in the feature space. Furthermore, Liang et al.~\cite{liang2021shotplus} design different self-supervised auxiliary prediction tasks.

However, prior works often assume that source and target domains share the same label space, heavily limiting the real-world applicability of TTA. 
In contrast, our method fully utilizes the knowledge learned by pre-trained vision language models and adapts it to any target domain.

\subsection{Vision-Language Models}
\label{sec:Vision-language Pre-training}
Vision-language models (VLMs) have made remarkable advances in recent years, achieving surprising progress in computer vision performance and various application scenarios, such as image generation~\cite{ramesh2022Dalle2}, video retrieval~\cite{luo2022clip4clip}, audio processing~\cite{guzhov2022audioclip}, image segmentation~\cite{Wang_2022_CVPR_CLIPseg}, object detection~\cite{gu2021open}, etc. VLMs provide innovative solutions for existing tasks and extend the potential application range of visual models~\cite{zhang2023_VLP_vision}.

VLMs map text and images into the same high-dimensional space, associating the ``concepts'' represented by text with images. This is typically achieved using a dual encoder structure that processes text and image data separately~\cite{gan2022_VLP_survey}. 
The success of VLMs can be attributed to developments in Transformers~\cite{devlin-etal-2019-bert}, contrastive learning~\cite{he2020_MoCo,SimCLR-chen20j,gan2022_VLP_survey}, and large-scale dataset~\cite{jia2021_ALIGN,schuhmann2022laion,gu2022wukong}. 
A representative model is CLIP~\cite{radford2021learning}, a multimodal pre-trained model that learns image-text correspondences through contrastive learning. CLIP has demonstrated remarkable zero-shot transferability in a wide range of downstream tasks, making it a natural choice as a general pre-trained model for TTA. 

VLMs are progressively replacing traditional supervised pre-training models, becoming the new foundational models for visual tasks~\cite{liu2024few}. VLMs demonstrate significant advantages in addressing fragmented and diverse challenges. Traditional models typically require training and optimization on specific datasets for designated tasks, limiting their generalizability and applicability to only single tasks, which leads to high overall costs. In contrast, VLMs offer more flexible solutions, providing a robust performance base for various downstream tasks.

\subsection{Parameter-efficient Fine-tuning}
\label{sec:PEFT}
Fine-tuning on pre-trained models has emerged as a fundamental paradigm in computer vision~\cite{ding2023parameter}. Models are typically pre-trained on large datasets and then fine-tuned for specific tasks, which often requires significant, high-quality annotated data and substantial computational resources. However, fully fine-tuning large models such as CLIP can be costly and impractical. 
In response, various strategies have been developed that only require fine-tuning a small number of parameters, with little or minimal data to achieve comparable or even superior performance to full fine-tuning~\cite{lialin2023scaling,ding2022delta}.
In this section, we focus on CLIP’s parameter-efficient fine-tuning.

One aspect of parameter-efficient fine-tuning is~\textit{fine-tuning an existing part} of CLIP. 
A typical method is prompt tuning. CoOp~\cite{zhou2022CoOp} and CoCoOp~\cite{zhou2022_CoCoOp} achieve automatic context learning for specific categories, and ProGrad~\cite{zhu2023_ProGrad} optimizes prompt updates using gradient consistency. Methods of this type require only a few high-quality labeled images per category to significantly enhance CLIP's performance on specific tasks.
Besides, Zaekn et al.~\cite{ben-zaken-etal-2022-bitfit} propose that fine-tuning only the bias terms is competitive with full fine-tuning. 

Another aspect is training an extra module, which usually contains a small number of parameters \cite{gao2021clip,yu2023task,li2024graphadapter,lu2023beyond}. 
CLIP-Adapter~\cite{gao2021clip} suggests inserting an adapter~\cite{houlsby2019parameter} following the image or text encoder of CLIP. Ding et al.~\cite{ding2023parameter} prepend trainable parameters to the hidden layers. In these models, only the added modules are trained from scratch, while other parts are frozen.

In this paper, we use CLIP as the pre-trained model to investigate the TTA task. 
Instead of fully fine-tuning the whole model, we add a CAM to extract the knowledge from the target domain while accelerating the fine-tuning process. Experimental results show that our method surpasses the prompt-tuning based methods such as CoOp~\cite{zhou2022CoOp}, CoCoOp~\cite{zhou2022_CoCoOp}, and TPT~\cite{shu2022TPT} (see Figure~\ref{fig:CLIP_Based}).

\section{Method}

Different from prior works addressing TTA, we propose a unified framework that could be adapted to an arbitrary target domain $\mathcal{D_T}$ without any information about source domain $\mathcal{D_S}$. 
Given a well-trained model $F(\theta )$, the purpose is to adapt it to unlabeled target data denoted as $\{x_{i}\}^n_{i=1} \in \mathcal{X_T}$. 
The corresponding labels $\{y_{i}\}^n_{i=1} \in \mathcal{Y_T}$ are exclusively for evaluation purposes. Generally, the prior information of label space $ \mathcal{R} $, which represents the set of all categories in $\mathcal{Y_T}$, is given. During training, pseudo-labels \(\hat{y}_i = F(\theta,x_{i}) \) are generated, filtered, and used to update the model. Exploiting knowledge of the source model implied in  $F(\theta )$, the module parameterized by $\delta$ is optimized to extract the target domain knowledge. The loss is formulated as follows:
\begin{equation}
\min_{L} \sum_{i=1}^{n} L(F(\theta, \delta, x_{i}, \mathcal{R}), \hat{y}_i).
 \label{eq:Definition}
\end{equation}

In the following, we first briefly introduce how CLIP is utilized for TTA. 
Then, we present the technical details of our approach, COLA, which includes CBPL and a CAM. 

\subsection{CLIP for TTA}

 CLIP utilizes an image encoder \(E_I\) and a text encoder \(E_T\) to map texts and images into a shared low-dimensional space. Zero-shot inference is implemented as follows:

Given the prior $ \mathcal{R} $, a set of prompts $\mathcal{P}=\{p_j\}_{j=1}^{c} $ can be constructed, where each $p_j$ is formatted as ``a photo of a \{classname\} and $c$ represents the class index.
Then, an image $x_i$ and a set of prompts $\mathcal{P}$ are processed to extract embedded features through \(E_I\) and \(E_T\), respectively, yielding \(f_i = E_I(x_i)\) and \(v_j= E_T(p_j)\). Subsequently, the similarity between $f_i$ and each element in the set $V=\{v_j\}_{j=1}^c$ is calculated to find the description that is closest to the image, thus obtaining the classification result. This is formally represented as:
\begin{equation}
p(p_j|x_i) = \frac{exp(f_i \cdot v_j / \tau)}{\sum_{m \in T} exp(f_i \cdot v_m / \tau)},
\label{eq:CLIP}
\end{equation}
where \(\tau\) is a temperature parameter that controls the sharpness of the softmax function. Upon acquiring the conditional probability \(p(p_j|x_i)\), the label of the input image is predicted by identifying the category with the highest probability, represented as: \(\hat{y}_i = argmax(\{p(p_j|x_i)|j \in \{1, \dots, c\}\})\).

CLIP’s inherent suitability for TTA stems from its image-text alignment classifier. 
This offers more flexibility compared to the fully connected classifiers used in previous TTA works, allowing adaptation to any target domain without being constrained by the shared label space between source and target domains.

\begin{figure*}[!t]
  \centering
  \includegraphics[width=\textwidth, trim={0cm 0.3cm 0cm 0cm},clip]{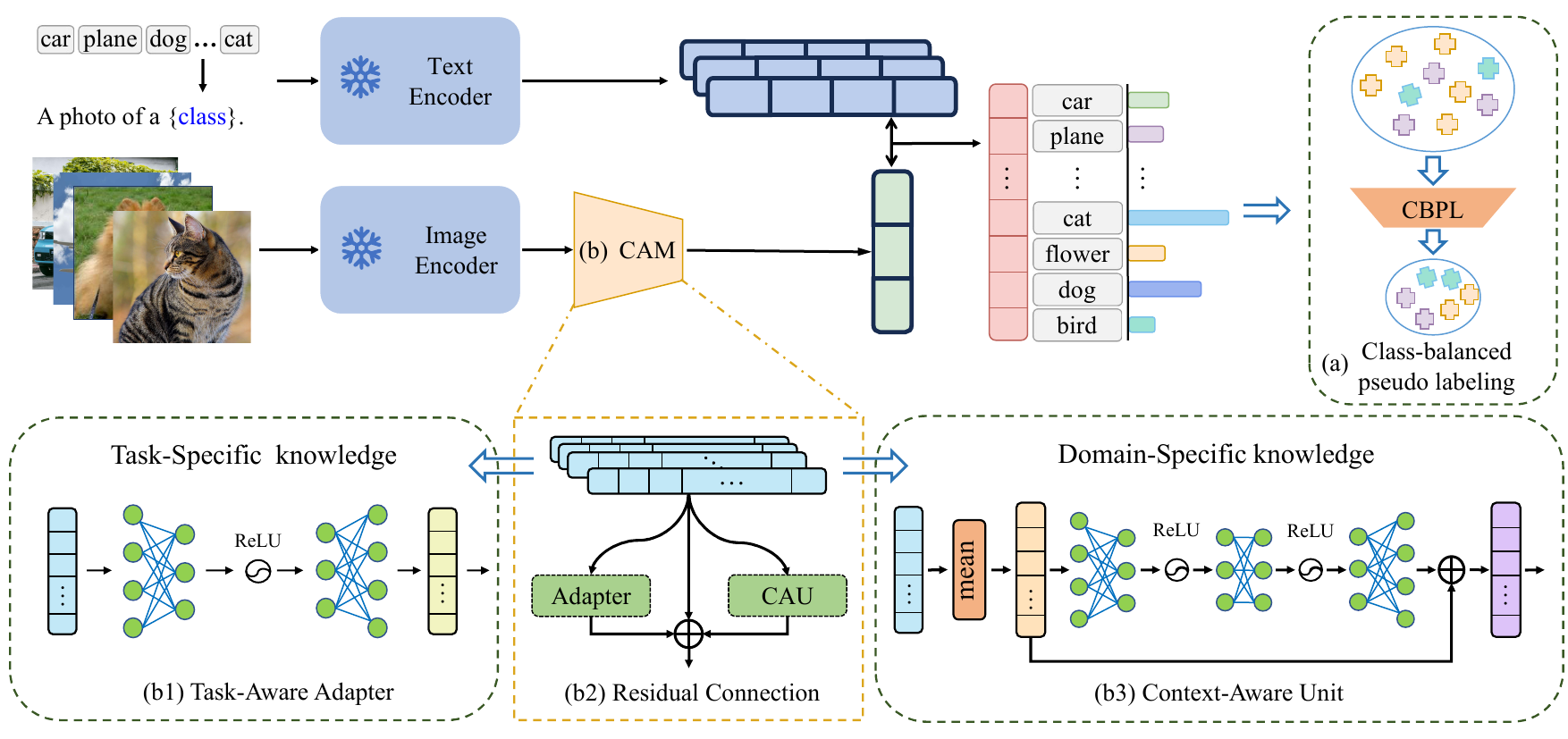}
  \caption{Framework of COLA. Based on CLIP, COLA consists of two key components to learn target domain knowledge while utilizing the inherent knowledge of the CLIP model. (a) To make full use of the knowledge learned by CLIP and the trade-off between quality and quantity of pseudo labels, we use CBPL to select high-confidence training samples for CAM, which includes the following components. (b1) Task-aware adapter acquires special knowledge from target samples. (b2) Context-aware unit (CAU) extracts domain-specific knowledge from the target domain by calculating domain-wise prototypes within batch-wise data. (b3) The operator $\oplus$ denotes residual connections, which output the final visual features. Finally, a cross-entropy loss is employed to update the parameters in CAM while keeping the image and text encoders of CLIP frozen (indicated by a snowflake). }
  \label{fig:overview}
\end{figure*}

\subsection{Context-Aware Language-driven TTA}
\label{cola}
Although CLIP shows great potential for TTA tasks in a zero-shot manner, it is hard to fully fine-tune the whole model due to the expensive training cost for each target domain. 
Furthermore, the domain information implied in the target data is not exploited in CLIP.
To address these challenges, we focus on adapting CLIP for the TTA task, integrating its inherent knowledge with task-specific knowledge. 

In this section, we first present a CBPL approach that constructs a balanced training dataset, effectively leveraging the knowledge embedded in the CLIP model.
Subsequently, a novel CAM is introduced to distil implicit knowledge from the target domain.

  \subsubsection{\textbf{Class-Balanced Pseudo-Labeling (CBPL)}}
\label{sec:CBPL}
In general, fixed-threshold pseudo-label acquisition retains high-confidence labels, reducing noise and error accumulation~\cite{chen2022_SSNLL}. However, domain shift may cause a trade-off between sample quality and quantity~\cite{saito2019semi}. 
A high threshold ensures quality but reduces the number of filtered samples, while a low value increases the number of samples, potentially compromising quality. Furthermore, a fixed threshold may lead to an overemphasis on ``simple'' category samples by the model, and ``difficult'' categories may not receive adequate attention, leading to a biased optimization due to imbalanced samples.

\begin{algorithm}[!t]
  \caption{Class-Balanced Pseudo-Labeling (CBPL)}
  \begin{algorithmic}[1]
    \Require unlabeled target data $\{x_i\}_{i=1}^n \in \mathcal{X}_T$, prompts $\mathcal{P}=\{p_j\}_{j=1}^C$, global confidence threshold $T_g$, retention ratio $Q$.
    \Ensure filtered dataset $\mathcal{D}' = \{(x'_i,\hat y'_i)\}$

    \State compute logits $L_{ij}=f(x_i, p_j;\theta)$ for all $i,j$
    \State obtain labels and confidences:
    \Statex \quad $\hat y_i=\arg\max_j softmax(L_{ij}), \quad c_i=argmax_j(L_{ij})$

    \For{$k=1$ \textbf{to} $C$}
      \State let $\mathcal{S}_k = \{(x_i,\hat{y}_i,c_i)\mid \hat y_i == k\}$. 
      \State sort $\mathcal{S}_k$ in descending order by $c_i$.
      \State $T_s \gets$ minimum confidence of top Q of samples.
      \State $t_k \gets \min(T_g, T_s)$
      \State $\mathcal{D}' \gets \{(x_i,\hat y_i)\mid c_i \ge t_{k}\}$
    \EndFor
    \State \Return $\mathcal{D}'$
  \end{algorithmic}
\end{algorithm}

To address this issue, we propose a pseudo-label filtering strategy that sets distinct thresholds for each class, balancing between sample quantity and quality. 
Specifically, we introduce two thresholds to manage the filtering process: a global confidence threshold \(T_g\) and a sample retention threshold \(T_s\). Initially, CLIP generates pseudo-labels for all samples. In order to ensure a sufficient representation of each class in the training dataset, we retain the top $Q$ of samples from each class based on confidence scores. 
Following this, the lowest confidence score among the selected samples in each class is set as the sample retention threshold \(T_s\). For each class $k$, the class-specific threshold \(t_k\) is the lesser of \(T_g\) and \(T_s\). We collect these into the set $\mathcal{T} = \{t_k\}_{k=1}^C$, with $C$ denoting the total number of classes.
Through this class-wise thresholding approach, our CBPL can effectively balance the quantity and quality of pseudo-labels.

 \subsubsection{\textbf{Context-aware Module (CAM)}}
 
We propose a plug-and-play CAM that is concatenated with the image encoder of CLIP. The CAM is designed to learn task-specific knowledge and leverage implicit domain-specific knowledge. It consists of a task-aware adapter, a CAU, and a residual connection unit. 
This enables the integration of the VLM’s prior knowledge and task-specific insights to work for downstream tasks.

\paragraph{Task-aware Adapter} 
Inspired by CLIP-Adapter, we use the ``task-aware adapter'' module, composed of two learnable linear layers, to learn task-specific knowledge from each sample in the target domain.
The output of the adapter \(f_i^{\text{A}}\) is combined with the original output \(f_i\) of the image encoder in CLIP. 
This design enables CLIP's adaptation to the target domain while preserving the original knowledge encoded in the pre-trained model.

\paragraph{Context-aware Unit (CAU)} is designed to leverage the prior knowledge embedded in VLMs to extract domain-specific knowledge from the target domain. Specifically, CAU comprises a multi-layer perceptron with residual connections that processes the batch-wise mean feature vector. The mean vector is used as an initial approximation of the domain distribution and is subsequently refined by the CAU to capture meaningful domain-specific knowledge. 
This process can be viewed as a form of meta-learning, enabling the CAU to learn how to extract meaningful domain-specific knowledge from the target domain.
Simultaneously, CAU associates and reuses highly correlated knowledge, which better aligns the semantics between the textual and visual representations in the embedding space.

Formally, the input to CAU is a batch of features ${\{f_i\}}_{i=1}^n$, extracted from the image encoder \( E_T\) of CLIP. The mean of these features is computed along the batch dimension as $\bar{f} = \frac{1}{n} \sum_{i=1}^n f_i$, representing the domain-wise prototypes within the batch. These mean features are then passed through a multi-layer perceptron (MLP) with residual connections. At inference time, computing the mean over a single input yields a feature with batch dimension 1, ensuring that CAU correctly handles inputs regardless of batch size.
The proportion of the residual is controlled by a scalable parameter $\lambda$, which is constrained by a sigmoid function $\sigma$. Therefore, the process of obtaining the CAU feature can be formalized as follows:
\begin{equation}
f_i^{\text{CAU}} = \text{MLP}(\bar{f}) + \sigma(\lambda) \cdot \bar{f}.
\label{eq:CAU-train}
\end{equation}

The CAM output, serving as the final visual feature in our method, is derived from the \textit{residual} fusion of the original visual features in CLIP, the features extracted by CAU, and the task-aware Adapter.
The final visual feature is as follows:

\begin{equation}
f_{\text{CAM}} = \alpha f_i + \beta f_i^{\text{A}} + \gamma f_i^{\text{CAU}}.
\label{eq:CLIP-CAM}
\end{equation}

\subsection{Training Pipeline}
\label{sec:Pipeline}
COLA employs a two-stage strategy to achieve adaptation to the target domain, which is introduced as follows:

\begin{itemize}[leftmargin=*]
    \item \textbf{Data preparation by CBPL:} By harnessing the prior information of label space $ \mathcal{R} $, we construct prompt set $T$, which is then fed into CLIP with images. For each image, CLIP generates its pseudo label with a confidence score. CBPL, as detailed in Sec.~\ref{cola}, leverages global confidence and the ratio of sample quantities across categories to set personalized confidence thresholds for each category's pseudo-labels. We select images that surpass the respective category-specific thresholds to form the final training dataset for CAM to learn the knowledge from the target domain.
    
    \item \textbf{CAM training:} After the CBPL method has successfully filtered the images and generated a corresponding prompt set, this combined data is fed into CLIP. As delineated in Eq.~\ref{eq:CLIP-CAM}, the model extracts high-dimensional visual features from the images. These visual features are then aligned with their corresponding text features to generate the final predictions. To optimize CAM, we minimize a cross-entropy loss function, which measures the discrepancy between the model's predictions and the pseudo-labels provided by the CBPL method.
\end{itemize}

\begin{table*}[!t]
\centering
\caption{Classification accuracies (\%) for TTA on VisDA-C. 
* denotes the zero-shot results. 
}
\resizebox{1.95\columnwidth}{!}{
\small
\begin{tabular}{l|c|cccccccccccc|c}
    \toprule
        Method& Backbone  & plane & bcycl & bus & car & horse & knife & mcycl & person & plant & sktbrd & train & truck & Avg. \\ 
        \midrule
        CLIP*~\cite{radford2021learning} & RN-101 & 98.1 & 83.7 & 90.8 & 74.2 & 97.3 & 85.9 & 95.2 & 69.3 & 82.0 & 90.0 & 92.5 & 61.1 & 85.0 \\
        SHOT~\cite{liang2020we} & RN-101  & 94.3 & 88.5 & 80.1 & 57.3 & 93.1 & 94.9 & 80.7 & 80.3 & 91.5 & 89.1 & 86.3 & 58.2 & 82.9 \\
        
        SHOT++~\cite{liang2021shotplus} & RN-101  &  97.7 &  88.4  &  90.2 & 86.3 & 97.9  &  \textbf{98.6}  &  92.9  &  84.1  &  97.1  &  92.2  &  93.6  &  28.8  &  87.3 \\

        NRC~\cite{yang2021_NRC}  & RN-101  &  96.8 & 91.3 & 82.4 & 62.4 & 96.2 & 95.9 & 86.1 & 80.6 & 94.8 & 94.1 & 90.4 & 59.7 & 85.9 \\
 
        CPGA~\cite{qiu2021_CPGA} & RN-101 &  95.6 & 89.0 & 75.4 & 64.9 & 91.7 & 97.5 & 89.7 & 83.8 & 93.9 & 93.4 & 87.7 & 69.0 & 86.0 \\

        VDM-DA~\cite{tian2021vdm}  & RN-101  &  96.9  &  89.1  &  79.1  &  66.5  &  95.7  &  96.8  &  85.4  &  83.3  &  96.0  &  86.6  &  89.5  &  56.3  &  85.1 \\
        
        CDCL~\cite{wang2022_CDCL}  & RN-101  &  97.4  &  89.5  &  85.9  &  78.2  &  96.4  &  96.8  &  91.4  &  83.7  &  96.3  &  96.2  &  89.7  &  61.6  &  88.6 \\

        AdaCon~\cite{Chen_2022_CVPR}  & RN-101  &  97.2 & 87.0 & 86.7 & 81.7 & 95.5 & 91.6 & 93.5 & \textbf{86.6} & 95.3 & 90.9 & 92.8 & 47.9 & 87.2 \\

        C-SFDA~\cite{karim2023_C-SFDA} & RN-101  & 97.6 & 88.8 & 86.1 & 72.2 & 97.2 & 94.4 & 92.1 & 84.7 & 93.0 & 90.7 & 93.1 & 63.5 & 87.8\\

        TPT~\cite{shu2022TPT} & RN-101 & 97.5 & 84.2 & 88.6 & 74.3 & 96.8 & 86.1 & 94.2 &  84.3 & 79.0 & 90.4 & 91.4 & 67.7 & 86.2 \\

        \rowcolor{Gray}
        COLA (Ours) & RN-101 &\textbf{99.7} & \textbf{94.6} & \textbf{98.4} & \textbf{89.6} & \textbf{99.5} & 95.5 & \textbf{99.9} & 80.1 & \textbf{97.6} & \textbf{98.3} & \textbf{98.6} & \textbf{81.9} & \textbf{94.5} 
         \\
        \midrule
        CLIP*~\cite{radford2021learning} & ViT-B/16 & 99.2 & 92.3 & 93.5 & 76.6 & 98.3 & 90.4 & 94.6 & 83.6 & 85.4 & 96.1 & 94.3 & 62.7 & 88.9\\

        TENT~\cite{wang2020tent} & ViT-B/16 & 99.0 & 76.9 & 79.2 & 80.8 & 93.9 & 84.2 & 95.9 & 54.5 & 74.6 & 92.9 & 95.6 & 22.9 & 79.2 \\
        
        SHOT~\cite{liang2020we} & ViT-B/16  &  99.5 & 91.8 & 88.7 & 65.1 & 98.6 & 98.0 & 96.0 & 66.1 & 95.1 & 98.9 & 96.8 & 52.4 & 87.3 \\
        
        AdaCon~\cite{Chen_2022_CVPR} & ViT-B/16 & 99.5 & 94.2 & 91.2 & 83.7 & 98.9 & 97.7 & 96.8 & 71.5 & 96.0 & 98.7 & 97.9 & 45.0 & 89.2 \\

        DePT~\cite{gao2022_DEPT} & ViT-B/16 & 99.4 & 93.8 & 94.4 & 87.5 & 99.4 & 98.0 & 96.7 & 74.3 & 98.4 & 98.5 & 96.6 & 51.0 & 90.7 \\ 
        
        TPT~\cite{shu2022TPT} & ViT-B/16 & 97.5 & 94.5  & 81.6 & 88.9  & 93.3 & 90.6 & 91.3 & 80.7 & \textbf{98.6} & 80.3 & 96.4 & 82.7 & 89.8 \\
        
        \rowcolor{Gray}
        COLA (Ours) & ViT-B/16  & \textbf{100} & \textbf{99.9} & \textbf{99.0} & \textbf{97.4} & \textbf{99.6} & \textbf{97.5} & \textbf{100} & \textbf{97.3} & 98.0 &  \textbf{99.9}  & \textbf{99.8} & \textbf{91.4} & \textbf{98.3} \\
        \bottomrule
    \end{tabular}
}

\label{tab:visda-c_main}
\end{table*}

\begin{table}[t]
    \small
    \centering
    \caption{Classification accuracies (\%) for TTA on DomainNet-126.}
    \resizebox{1\columnwidth}{!}{
    \begin{tabular}{l|c|cccc|c}
    \toprule
        Method & Backbone & R & C & S & P & Avg. \\ \midrule
        CLIP*~\cite{radford2021learning} & RN-101 & 85.2 & 	73.7 & 72.3 & 73.2 & 76.1 \\
        SHOT~\cite{liang2020we}  & RN-101 & 83.3 & 77.2 & 67.8& 72.7 & 75.3\\
        AdaCon~\cite{Chen_2022_CVPR}& RN-101 & 83.9 & \textbf{80.1} & 71.5 & 75.2 & 77.7 \\
        
        \rowcolor{Gray}
        COLA (Ours) & RN-101 & \textbf{90.1} & 79.9 & \textbf{75.6} & \textbf{78.8} & \textbf{81.1} \\
        \midrule
        CLIP*~\cite{radford2021learning} & ViT-B/16  & 92.0 & 83.9 & 81.0 & 82.3 & 84.8 \\
        SHOT~\cite{liang2020we} & ViT-B/16 & 90.7 & 83.1 & 75.9 & 83.4 & 83.2 \\ 
        AdaCon~\cite{Chen_2022_CVPR} & ViT-B/16 & 85.4 & 79.6 & 71.1 & 75.9 & 78.0 \\
        DePT~\cite{gao2022_DEPT} & ViT-B/16 & 91.0 & 83.7 & 78.3 & 84.0 & 84.2 \\
        \rowcolor{Gray}
        COLA (Ours) & ViT-B/16 & \textbf{93.1} & \textbf{86.5} & \textbf{82.9} & \textbf{85.5} & \textbf{87.0}
        \\

        \bottomrule
    \end{tabular}
    }
    \label{tab:DomainNet126_main}
\end{table}

\section{Experiments}

Our method is mainly evaluated in two setups: (i) test-time adaptation on five major benchmarks; (ii) base-to-new generalization within VisDA-C. In the following, we first provide the details of the experimental setup. 
Then, we compare COLA with the previous SOTA methods on TTA and base-to-new generalization tasks, followed by further analysis, including comparison with CLIP-based methods, low-shot learning, efficiency, ablation study, and visualization.

\subsection{Experimental Setup}

\subsubsection{Datasets} Experiments are conducted on the following popular domain adaptation datasets. Note that our method can adapt to any target domain; therefore, only the validation sets of these datasets are used.
\begin{itemize}[leftmargin=*]
    \item{VisDA-C~\cite{peng2017visda}} is a large-scale domain adaptation benchmark that comprises 12 categories with over 50k real images in its validation set.
    \item{DomainNet-126~\cite{saito2019semi}} is a subset of DomainNet with four domains: Real(R), Clipart(C), Sketch(S), and Painting(P), each with 126 unique categories, totalling over 140K images. The noise, diversity of categories, and disparity in samples make the dataset harder to learn.
    \item{VLCS~\cite{khosla2012undoing_VLCS}} consists of images selected from four widely used image classification datasets—Caltech101 (V), LabelMe (L), SUN09 (C), and VOC2007 (S), each serving as a distinct domain.
    The number of samples within each category varies, ranging from 20 to 1,499.
    \item{PACS~\cite{li2017deeper_PACS}} includes four domains characterized by significant shifts and diverse visual styles: photos, art paintings, cartoons, and sketches. 
    It encompasses seven categories within each domain, providing a rich variety for cross-domain studies.
    \item{OfficeHome~\cite{venkateswara2017_OfficeHome}} is a domain adaptation benchmark dataset that encompasses four domains: Art, Clipart, Product, and Real-World, with 65 categories of common items, each category averaging 70 samples.%
\end{itemize}

\subsubsection{Evaluation Metrics}

To thoroughly validate the adaptation capability and generalization of our proposed method, we conduct two types of experiments: test-time adaptation and base-to-new. 
In the TTA experiments, we assess the model performance by top-1 accuracy and average (Avg.). For VISDA-C, top-1 accuracies for each class are reported. Similarly, for other datasets, top-1 accuracies for each domain are also reported.
For the base-to-new experiments, we evaluate the model's generalization by reporting the accuracy of both the base classes and the novel classes, and the harmonic mean of these accuracies~\cite{xian2017zero_harmonic}.

\subsubsection{Baselines} 
We compare our method with representative TTA methods, including pseudo-labeling methods: SHOT~\cite{liang2020we}, C-SFDA~\cite{karim2023_C-SFDA} and NRC~\cite{yang2021_NRC}, consistency methods: AdaCon~\cite{Chen_2022_CVPR}, DePT~\cite{gao2022_DEPT}, clustering methods: CDCL~\cite{wang2022_CDCL}, source estimation methods: VDM-DA~\cite{tian2021vdm} and CPGA~\cite{qiu2021_CPGA}, entropy minimization methods: TENT~\cite{wang2020tent}, self-supervision methods: SHOT++~\cite{liang2021shotplus}, and CLIP-based methods: TPT~\cite{shu2022TPT}. 
For base-to-new generalization tasks, we compare with CLIP-based methods, including CoOp~\cite{zhou2022CoOp} and CoCoOp~\cite{zhou2022_CoCoOp}.

\subsubsection{Implementation Details}
We consider CLIP ResNet-101 (RN-101) and ViT-B/16 as the base network for all implementations. The target domain training is conducted over 15 epochs for all datasets. The SGD optimizer is employed with a momentum of 0.9 and a weight decay of 1e-6, and the learning rate is adjusted using a cosine annealing strategy. The batch size is set to 128.
In CBPL, we uniformly set the value of the global threshold \(T_p\) to 0.75, while the value of \(Q\) is set to 0.75 on VISDA-C and 0.2 on DomainNet.
The number of neurons in the hidden layer of the CAU and the adapter in CAM are both set to one-fourth of the input layer.

\begin{table}[!t]
    \centering
    \setlength{\tabcolsep}{6pt}
    \caption{Classification accuracies (\%) for TTA on VLCS.} 
    \begin{tabular}{l|c|cccc|c}
    \toprule
     method & Backbone & V & L & C & S & Avg. \\
         \midrule
         CLIP*~\cite{gao2021clip}  & RN-101 & 77.2 & 60.1 & 99.7 & 67.9 & 76.2 \\
         SHOT~\cite{liang2020we}   & RN-101 & 77.2 & 62.2 & 98.7 & 68.5 & 76.7 \\
         TPT~\cite{shu2022TPT}   & RN-101 & 80.1 & 64.9 & 96.2 & 72.8 & 78.5 \\
         AdaCon~\cite{Chen_2022_CVPR}   & RN-101 & 81.2 & 69.8 & 96.9 & 74.7 & 80.7 \\
         \rowcolor{Gray}
         COLA (Ours) & RN-101 & \textbf{85.5} & \textbf{70.7} & \textbf{100} & \textbf{83.1} & \textbf{84.8}\\
         \midrule
         CLIP*~\cite{gao2021clip}  & ViT-B/16 & 85.9 & 69.2 & 99.9 & 75.3 & 82.6 \\
         SHOT~\cite{liang2020we}   & ViT-B/16 & 70.8 & 67.7 & 97.0 & 74.5 & 77.5 \\
         TPT~\cite{shu2022TPT}   & ViT-B/16 & 86.1 & 67.5 & \textbf{100} & 78.9 & 83.1\\
         AdaCon~\cite{Chen_2022_CVPR} & ViT-B/16 & 82.6 & 69.5 & 97.4 & 83.5 & 83.3 \\
        \rowcolor{Gray}
         COLA (Ours)  & ViT-B/16 & \textbf{92.3} & \textbf{77.9} & \textbf{100} & \textbf{87.8} & \textbf{89.5} \\
    \bottomrule
    \end{tabular}
    \label{tab:Supp_VLCS}
\end{table}

\subsection{Test-time adaptation}

\paragraph{VisDA-C} The experimental results on the VisDA-C dataset are shown in Table~\ref{tab:visda-c_main}. 
It is evident that our analysis reveals several key findings. 
(i) COLA demonstrates superior performance on both RN-101 and ViT-B/16 backbones, surpassing previous methods. With RN-101, COLA achieves the highest average accuracy, with an improvement of 5.9\% over the best prior method. On the ViT-B/16 architecture, it secures the highest accuracy in ten out of twelve categories, improving average accuracy by 7.6\% beyond the previous SOTA. 
(ii) CLIP exhibits significant promise for the TTA task, owing to its versatile adaptability to a wide array of target domains and its exceptional discriminative power across various classes. CLIP's superior performance is particularly evident in the ``person'' and ``truck'' categories when employed with a ViT-B/16 backbone.
(iii) Though CLIP performs well, it still falls short compared with TTA-specific methods. With CBPL and CAM introduced in Sec.~\ref{cola}, our method performs much better. Specifically, COLA results in more than a 10.0\% performance increase in five categories using RN-101, particularly with gains of 15.6\% in ``plant'' and 20.8\% in ``truck'', representing an average increase of 9.5\% over CLIP. 
For ViT-B/16, the model shows substantial improvements in four categories, exceeding 10\% in each. Notably, the ``car'' and ``truck'' categories achieve remarkable enhancements, with increases of 20.8\% and 28.7\%, respectively. The overall improvement over CLIP for ViT-B/16 averages 9.4\%.

The main reason behind our method's success lies in its ability to not only leverage task-specific knowledge but also extend its exploration to the domain-specific knowledge obtained from CLIP. This combination of expertise allows COLA to achieve remarkable performance gains.

\paragraph{DomainNet-126} Our method, \modelAbbr, outperforms current TTA methods on DomainNet-126, as shown in Table~\ref{tab:DomainNet126_main}. Similar conclusions can be drawn. COLA achieves the highest per-domain accuracy among all existing TTA methods and attains an impressive average accuracy of 87.0\%, surpassing CLIP and other SOTA methods by a significant margin of 3.4\% and 2.2\% with RN-101 and ViT-B/16, respectively. However, we find that the zero-shot results of CLIP with ViT-B/16 marginally surpass the previous SOTA. One possible reason is that the prior TTA methods struggle with adapting to the target domain due to large class sizes and intra-class variability on DomainNet-126.

\begin{table}[t]
    \centering
    \small
    \setlength{\tabcolsep}{3pt}
    \caption{Classification accuracies (\%) for TTA on PACS.}
    \begin{tabular}{l|c|cccc|c}
    \toprule
     method & Backbone & Photo & Art & Cartoon & Sketch & Avg. \\
         \midrule
         CLIP*~\cite{gao2021clip}  & RN-101 & \textbf{99.7} & 95.1 & 96.8 & 87.4 & 94.8 \\
         SHOT~\cite{liang2020we}   & RN-101 & 98.8 & 91.6 & 89.4 & 76.2 & 89.0\\
         TPT~\cite{shu2022TPT}   & RN-101 & 99.7 & 95.9 & 96.9 & 88.2 & 95.2 \\
         AdaCon~\cite{Chen_2022_CVPR} & RN-101 & 99.4 & 95.0 & 90.9 & 88.1 & 93.4\\
         
         \rowcolor{Gray}
         COLA (Ours) & RN-101 & \textbf{99.7} & \textbf{96.7} & \textbf{99.1} & \textbf{93.4} & \textbf{97.2} \\
         \midrule
         CLIP*~\cite{gao2021clip}  & ViT-B/16 & \textbf{99.9} & 97.2 & 99.1 & 88.2 & 96.1 \\
         SHOT~\cite{liang2020we}   & ViT-B/16 & 99.1 & 92.5 & 86.3 & 78.0 & 89.0 \\
         TPT~\cite{shu2022TPT}    & ViT-B/16 & \textbf{99.9} & 98.0 & 99.0 & 89.0 & 96.5\\
         AdaCon~\cite{Chen_2022_CVPR} & ViT-B/16 & 99.5 & 95.3 & 81.8 & 85.1 & 90.4 \\
        \rowcolor{Gray}
        COLA (Ours)   & ViT-B/16 & 99.7 & \textbf{99.0} & \textbf{99.6} & \textbf{93.7} & \textbf{98.0} \\ 
    \bottomrule
    \end{tabular}
    \label{tab:Supp_PACS}
\end{table}

\paragraph{VLCS}
As shown in Table~\ref{tab:Supp_VLCS}, an analysis of the detailed results of the VLCS dataset reveals the following key findings: (i) COLA achieves the best average accuracy on backbones such as RN-101 and ViT-B/16. Specifically, it improves performance by 4.1\% over AdaCon on RN-101, and by 6.2\% on ViT-B/16. (ii) COLA effectively harnesses the potential of large-scale vision-and-language pre-trained models. For example, despite the lower zero-shot performance of the CLIP baseline with RN-101, COLA still manages an 8.6\% improvement in performance on this backbone. Similarly, on the ViT-B/16 backbone, COLA achieves a 6.9\% improvement. (iii) Without adaptation, CLIP's zero-shot performance already shows strong generalizability and classification capabilities across various domains, indicating its substantial potential for test-time adaptation.

\begin{table}[t]
    \centering
    \small
    \setlength{\tabcolsep}{3pt}
    \caption{Classification accuracy (\%) on OfficeHome.}
    \begin{tabular}{l|c|cccc|c}
    \toprule
     method & Backbone & Art & Clipart & Product & Real & Avg. \\
         \midrule
         CLIP*~\cite{gao2021clip}  & RN-101 & 77.1 & 56.3 & 86.0 & 85.2 & 76.2 \\
         SHOT~\cite{liang2020we}   & RN-101 & 77.4 & \textbf{65.8} & 85.9 & 85.8 & 78.7 \\
         TPT~\cite{shu2022TPT}   & RN-101 & \textbf{79.8} & 56.5 & 85.9 & 86.2 & 77.1 \\
         AdaCon~\cite{Chen_2022_CVPR} & RN-101 & 72.2 & 63.1 & 84.8 & 82.7 & 75.7 \\
         \rowcolor{Gray}
         COLA (Ours)  & RN-101 & 78.3 & 60.5 & \textbf{89.9} & \textbf{87.4} & \textbf{79.0} \\
         \midrule
         CLIP*~\cite{gao2021clip}  & ViT-B/16 & 82.3 & 66.9 & 91.3 & 89.7 & 82.6 \\
         SHOT~\cite{liang2020we}   & ViT-B/16 & 78.5 & 62.5 & 85.7 & 85.5 & 78.1 \\
         TPT~\cite{shu2022TPT}   & ViT-B/16 & 84.3 & 68.8 & 89.6 & 90.4 & 83.3 \\
         AdaCon~\cite{Chen_2022_CVPR} & ViT-B/16 & 77.1 & 64.3 & 87.9 & 87.3 & 79.2 \\
        \rowcolor{Gray}
         COLA (Ours)   & ViT-B/16 & \textbf{84.8} & \textbf{71.0} & \textbf{93.2} & \textbf{91.5} & \textbf{85.1} \\
    \bottomrule
    \end{tabular}
    \label{tab:Supp_OfficeHome}
\end{table}

\paragraph{PACS}
Table~\ref{tab:Supp_PACS} presents the detailed results of the PACS dataset. The findings indicate the following: (i) The COLA method consistently outperforms other approaches in all tested architectures, including RN-101 and ViT-B/16. Specifically, it achieves an average performance increase of 2.4\% over the baseline and 2.0\% over TPT on RN-101. Similarly, on ViT-B/16, it surpasses the baseline by 1.9\% and TPT by a significant margin of 1.5\%. (ii) All methods show excellent performance in the PACS dataset, attributed to the fewer categories in the dataset, the large number of samples per category, and the significant variability between classes, which collectively reduce the learning challenge. Compared to prior methods, COLA not only leverages task-specific knowledge and effectively utilizes the priors from CLIP but also integrates domain-specific knowledge, thereby enhancing performance.

\paragraph{OfficeHome}
Detailed results on the OfficeHome dataset are shown in Table~\ref{tab:Supp_OfficeHome}. Given the OfficeHome dataset's wide variety of categories, each with a relatively small number of samples, and its complex scenarios, few papers have considered adapting larger models because such conditions make these models difficult to optimize. It’s worth noting that images in the Clipart domain of the OfficeHome dataset typically exhibit a distinctive style, representing various objects and scenes in a simplified and symbolic manner, thereby increasing the challenge of adaptation. Despite these difficulties, COLA improves performance by 2.8\% compared to the RN-101 baseline and by 2.5\% compared to the ViT-B/16 baseline.

\subsection{Base-to-new Generalization}
Inspired by CoCoOp, we conduct an experiment to assess the generalisation capability of our proposed method under the TTA framework.
Firstly, we split the dataset into base and new groups, each including six categories of unlabeled samples. We then train the model on the base group. After the training phase, we test the model on both base and new groups to evaluate its generalisability. Notably, instead of using ground truth labels, we use CLIP to obtain pseudo-labels for the base group to update the model. The results are shown in Table~\ref{tab:3a_VisDA-C_base_to_new}.
We formed six distinct groups:
\begin{itemize}[leftmargin=*]
    \item Group 1: Arranged in the order of dataset categories, with the initial six as base and the subsequent six as new.
    \item Group 2: Categorized based on category labels, with even numbers as base and odd numbers as new.
    \item Group 3: Arranged by the zero-shot accuracy of CLIP, with the group having the highest accuracy as base and the rest as new.
    \item Group 4-6: Control groups where we swapped the classes of base and new in groups 1-3.
\end{itemize}

As observed in Table \ref{tab:3a_VisDA-C_base_to_new}, COLA not only improves generalisation for new classes but also retains the base knowledge that is already learned.
Considering both base and new classes, COLA outperforms CoOp by over 4.9\%, demonstrating the effectiveness of the domain-specific knowledge extracted by our CAU.

\begin{table}[t]
\centering
\small
\setlength{\tabcolsep}{3pt}
\caption{ Classification accuracies (\%) 
for the base-to-new task on VisDA-C. H denotes Harmonic mean. Group 4-6, as control groups, swap the base and new classes in group 1-3. }
\small
\begin{tabular}{l|ccc|ccc}
\toprule
 ~ &  & Group1 &  &  & Group4 &  \\
 \cline{2-7}
  & \raisebox{-0.4ex}{Base} & \raisebox{-0.4ex}{New} & \raisebox{-0.4ex}{H} & \raisebox{-0.4ex}{Base} & \raisebox{-0.4ex}{New} & \raisebox{-0.4ex}{H} \\ 
 \midrule
 CLIP*~\cite{radford2021learning} &  96.2 & 92.6 & 94.4 & 92.6 & 96.2 & 94.4 \\
 CoOp~\cite{zhou2022CoOp}  &  96.9 & 92.7 & 94.8 & 94.2 & 96.2 & 95.2 \\
 CoCoOp~\cite{zhou2022_CoCoOp} &  96.7 & 86.8 & 91.5 & 93.6 & 96.3 & 94.9 \\
 \rowcolor{Gray}
 COLA (Ours)&    \textbf{99.5} & \textbf{97.7} & \textbf{98.6} & \textbf{98.2} & \textbf{99.4} & \textbf{98.8} \\
 \midrule
 ~ &  & Group2 &  &  & Group5 &  \\
 \cline{2-7}
  & \raisebox{-0.4ex}{Base} & \raisebox{-0.4ex}{New} & \raisebox{-0.4ex}{H} & \raisebox{-0.4ex}{Base} & \raisebox{-0.4ex}{New} & \raisebox{-0.4ex}{H} \\ 
 \midrule
 CLIP*~\cite{radford2021learning}& 96.9 & 88.0 & 92.2 & 88.0 & 96.9 & 92.2\\
 CoOp~\cite{zhou2022CoOp} & 97.9 & 86.3 & 91.7 & 89.7 & 97.2 & 93.3  \\
 CoCoOp~\cite{zhou2022_CoCoOp}& 97.8 & 86.9 & 92.0 & 88.6 & 97.2 & 92.7\\
 \rowcolor{Gray}
 COLA (Ours)& \textbf{99.6} & \textbf{96.0} & \textbf{97.8} & \textbf{95.6} & \textbf{99.4} & \textbf{97.5}\\
 \midrule
 ~ &  & Group3 &  &  & Group6 &  \\
 \cline{2-7}
  & \raisebox{-0.4ex}{Base} & \raisebox{-0.4ex}{New} & \raisebox{-0.4ex}{H} & \raisebox{-0.4ex}{Base} & \raisebox{-0.4ex}{New} & \raisebox{-0.4ex}{H} \\ 
 \midrule
 CLIP*~\cite{radford2021learning} & 97.5 & 86.6 & 91.7 & 86.6 & 97.5 & 91.7\\
 CoOp~\cite{zhou2022CoOp} & 97.7 & 84.9 & 90.9 & 89.0 & 97.1 & 92.9 \\
 CoCoOp~\cite{zhou2022_CoCoOp} & 97.4 & 85.6 & 91.1 & 88.1 & 97.1 & 92.4 \\
 \rowcolor{Gray}
 COLA (Ours)& \textbf{99.7} & \textbf{95.3} & \textbf{97.5} & \textbf{95.3} & \textbf{99.7} & \textbf{97.5} \\
 \bottomrule
\end{tabular}
    \label{tab:3a_VisDA-C_base_to_new}
\end{table}

\subsection{Further Analysis}

 \subsubsection{\textbf{Comparison with CLIP-based Methods}}

\paragraph{Setting} Although the backbones of TTA methods shown in Table~\ref{tab:visda-c_main} and the CLIP backbone both have the names RN-101 and ViT-B/16, minor differences in architecture and pre-trained weights significantly affect the TTA results. In this section, we validate the effectiveness of COLA by making a more fair comparison among the methods using the CLIP pre-trained image encoder. We compare with CLIP-based methods including CoOp~\cite{zhou2022CoOp}, CoCoOp~\cite{zhou2022_CoCoOp}, TPT~\cite{shu2022TPT} and a prior TTA SOTA method: AdaCon~\cite{Chen_2022_CVPR}.
For the AdaCon with CLIP backbones, we fine-tune only the image encoder of CLIP while the textual encoder of CLIP is kept frozen to maintain feature space alignment.

\begin{figure}[!t]
  \centering
  \includegraphics[width=0.5\textwidth]
  {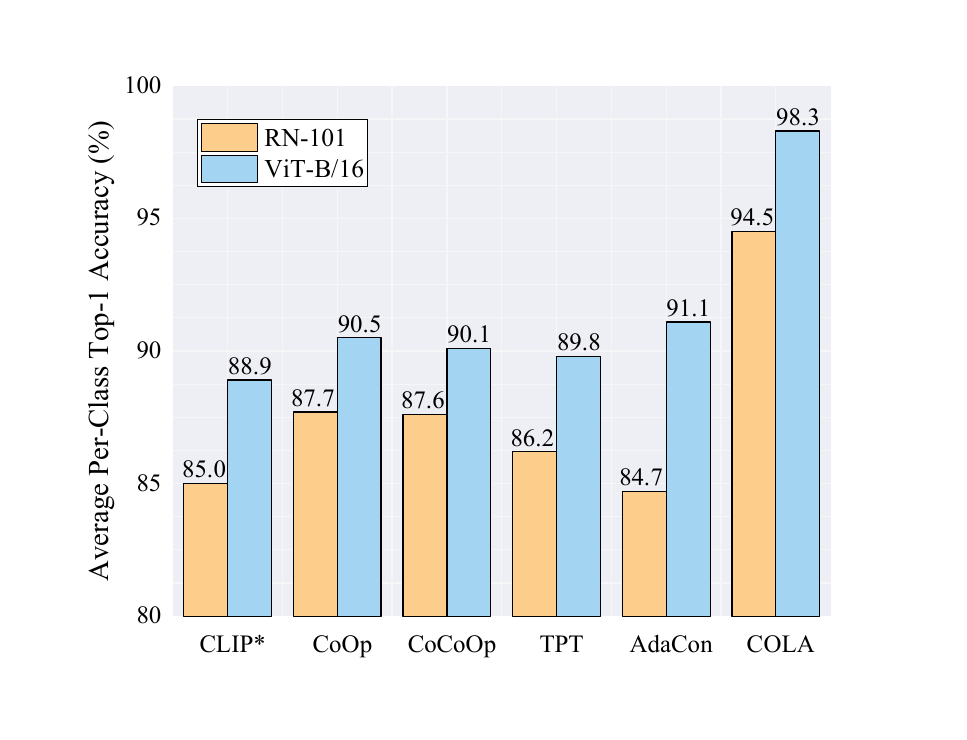}
  \caption{Comparison among different CLIP-based methods on VisDA-C. }
  \label{fig:CLIP_Based}
\end{figure}

\paragraph{Results} 
Figure \ref{fig:CLIP_Based} presents a comparative analysis of various CLIP-based methods on the VisDA-C dataset, revealing several key insights:
(i) Our proposed COLA method demonstrates a significant advancement over other CLIP-based methods. It achieves a remarkable improvement of 6.8\% and 7.2\% over the previous SOTA methods that utilize RN-101 and ViT-B/16 backbones, respectively. This substantial leap in performance underscores the effectiveness of COLA in adapting to the target domain.
(ii) The performance of prompt tuning methods, such as CoOp and CoCoOp, is observed to be superior to the baseline CLIP models for the TTA task. This indicates that refining the existing components of CLIP can unlock greater accuracy. In contrast, COLA takes a different approach by incorporating the CAM, which is designed to leverage both task-specific and domain-specific knowledge, leading to enhanced outcomes.
(iii) An examination of the AdaCon results, as presented in Table~\ref{tab:visda-c_main} and Figure~\ref{fig:CLIP_Based}, reveals variations in performance depending on the backbone used.
Specifically, the AdaCon variant utilizing the CLIP model with an RN-101 backbone exhibits a 2.5\% decrease in performance, while the version with a ViT-B/16 backbone demonstrates a 1.9\% improvement. 
This suggests that the RN-101 backbone may not be as effective in capturing discriminative and transferable knowledge, a hypothesis supported by the comparatively weaker results of other RN-101-based methods.

  \subsubsection{\textbf{Low-shot Learning}}

\paragraph{Setting} 
We conduct low-shot experiments on the following classification datasets: 
\begin{itemize}[leftmargin=*]
    \item{Caltech101~\cite{fei2004learning_Caltech101}} Caltech101 is a well-known dataset containing images of 101 daily life object categories, including animals, transportation, appliances, instruments, and sports equipment. Each category has around 40 to 800 images, with most averaging about 50.
    \item{OxfordPets~\cite{parkhi2012cats_Pets}} referred to as \textit{Pets} in this paper, is a fine-grained dataset of pet images, comprising 25 dog breeds and 12 cat breeds, totalling 37 common household pets. Each category contains around 200 images captured in real-world environments.
    \item{OxfordFlowers (Flowers102)~\cite{nilsback2008automated_Flowers102}} includes 102 flower species found in the UK, with each category containing between 40 and 258 images. Some categories exhibit significant intra-class variations, while others display minimal inter-class differences.
    \item{UCF101~\cite{taori2020measuring_UCF101}} constructed from real videos on YouTube, is an action recognition dataset with 101 categories. It includes significant variations in equipment, target appearance, scale, background, and lighting. 
\end{itemize}

\begin{table*}[!t]
  \centering
  \caption{Comparison of computational cost and performance of TTA methods on VisDA-C.}
  \label{tab:tta_efficiency}
  \small
  \resizebox{2\columnwidth}{!}{
  \begin{tabular}{l|c|c|c|c|c|c|c}
    \toprule
    Model & Backbone & Params (Tot / Trainable) & FLOPs (G) & Act. Mem (GB) & Infer. Time (ms) & Throughput (img/s) & Accuracy (\%) \\
    \midrule
    CLIP  & RN-101  & 119.69M~/~-~~~~ & 64.81 & - & 1.24 & 809.55 & 85.00\\
    TENT    & RN-101  & 44.55M~/~105.34K   & 47.24 & 15.16 & 2.81 & 355.86 & 70.70 \\
    AdaCon  & RN-101  & 86.06M~/~43.03M   & 7.87 & 15.80 & 1.10 & 912.93 & 87.20 \\
    TPT  & RN-101  & 119.69M~/~2.05K   & 1171.28 & 0.35 & 96.91  & 10.34   & 86.20 \\
    COLA & RN-101  & 120.15M~/~0.46M & 64.81 & 0.69 & 1.28 & 778.87 & 94.50 \\
    \midrule
    CLIP  & ViT-B/16  & 149.62M~/~- & 69.05 & - & 1.87 & 536.23 & 88.90\\
    TENT & ViT-B/16  & 86.57M~/~38.40K   & 67.72 & 11.13 & 3.76 & 266.01 & 79.20 \\
    AdaCon & ViT-B/16 & 173.14M~/~86.57M   & 18.29 & 18.97   & 1.66  & 601.58    & 89.20 \\
    TPT & ViT-B/16 & 149.62M~/~2.05K   & 17676.07 & 0.31 & 107.60  & 9.30    & 89.80 \\
    COLA & ViT-B/16 & 150.08M~/~0.46M   & 69.05 & 0.62   & 1.88  & 532.69    & 98.30 \\
    \bottomrule
  \end{tabular}
  }
\end{table*}

\begin{figure}[!t]
  \centering
  \includegraphics[width=0.5\textwidth]{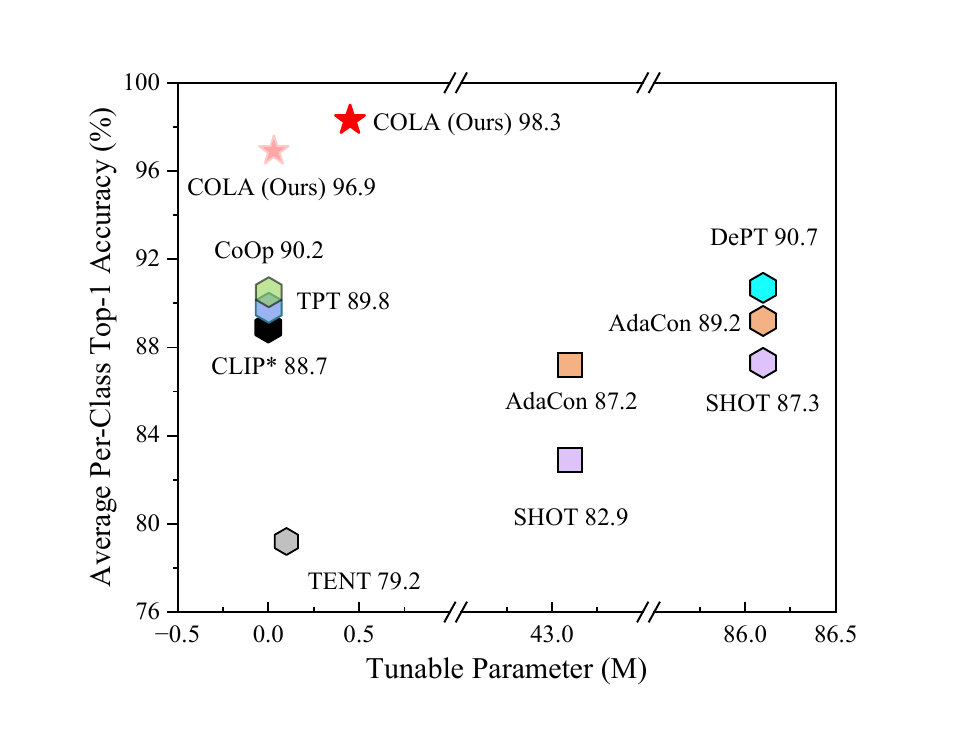}
  \caption{Comparison of TTA methods on VisDA-C. Rectangles denote methods with RN-101 backbone, while hexagons and stars use ViT-B/16. The star, COLA, achieves SOTA with few tunable parameters, achieving 98.3\% and 96.9\% with 0.03M and 0.45M parameters, respectively.}
  \label{fig:Efficiency}
\end{figure}

We further evaluate our method under 16-shot setup and compare with two SOTA low-shot learning based methods, \ieno, CoOp, and CoCoOp. 
We implemented COLA in CoOp’s official codebase and conducted 16-shot experiments as recommended by CoOp.
\paragraph{Results} 
The results, detailed in Table~\ref{tab:C_fewshot}, indicate that COLA markedly outperforms CoOp and CoCoOp in the low-shot setting over all four benchmarks. 
Utilizing the RN-101 backbone, our method yields an improvement of 2.6\%, 3.0\%, 0.9\%, and 6.1\% over other methods on Caltech101, OxfordPets, Flowers102, and UCF101, respectively.
With the ViT-B/16 backbone, the average performance enhancement is 14.1\% over CLIP's zero-shot performance and 2.3\% over CoOp. 
Note that CoCoOp focuses on generalization, and its performance gain in the low-shot scenario is marginal.

 \subsubsection{\textbf{Efficiency}}
To provide a comprehensive analysis of COLA’s efficiency, we first examine its parameter usage and then evaluate the computational costs of adapting COLA across different backbone networks. All experiments are conducted on the VisDA-C dataset using a single NVIDIA RTX 3090 GPU (24~GB). FLOPs are measured per image using the open-source tool THOP~\cite{zhu2022thop}. Activation memory, inference latency, and throughput are evaluated with a batch size of 128, except for AdaCon, which uses a batch size of 64 due to memory limitations. For TPT, we follow the official setting that adopts 63 test-time augmented views.

\begin{table}[!t]
    \centering
    \setlength{\tabcolsep}{2pt} 
    \small
    \caption{16-shot classification accuracies on Caltech101, Pets (OxfordPets), Flowers102, and UCF101.}
    \begin{tabular}{l|c|cccc}
    \toprule
     Method & Backbone & Caltech101 & Pets & Flowers102 & UCF101  \\ 
         \midrule
         CLIP*~\cite{radford2021learning}   & RN-101 & 89.7 & 86.9
 & 64.2 & 61.0 \\ 
         CoOp~\cite{zhou2022CoOp}   & RN-101 & 91.8 & 88.7
 & 95.2 & 78.0 \\ 
         CoCoOp~\cite{zhou2022_CoCoOp} & RN-101 & 92.9 & 91.2 & 83.4 & 72.5 \\ 
         
         \rowcolor{Gray}
         COLA (Ours)  & RN-101 & \textbf{95.5} & \textbf{94.2} & \textbf{96.1} & \textbf{84.1} \\ 
         \midrule
         CLIP*~\cite{radford2021learning}   & ViT-B/16 & 92.9 & 89.1
 & 71.2 & 66.7 \\ 
         CoOp~\cite{zhou2022CoOp}   & ViT-B/16 & 95.8 & 92.5 & 96.5 & 82.2 \\ 
         CoCoOp~\cite{zhou2022_CoCoOp} & ViT-B/16 & 95.1 & 93.6 & 89.1 & 77.1 \\ 
        \rowcolor{Gray}
         COLA (Ours)   & ViT-B/16 & \textbf{96.3} & \textbf{95.9} & \textbf{98.1} & \textbf{86.1} \\ 
    \bottomrule
    \end{tabular}
    \label{tab:C_fewshot}
\end{table}

COLA is designed to enable efficient test-time adaptation by incorporating a lightweight module (CAM) and a pseudo-label filtering strategy (CBPL) while keeping the CLIP backbone frozen. CAM is the only trainable component during adaptation, comprising two linear layers in the task-aware adapter and three in the CAU. To assess parameter efficiency, we compare COLA with representative TTA methods on VisDA-C. As shown in Figure~\ref{fig:Efficiency}, COLA achieves strong performance with relatively few trainable parameters. Notably, even when compressing CAM by reducing its hidden dimension to 16, COLA reaches 96.9\% accuracy with only 0.03M parameters, achieving superior performance to all baselines with similar or lower parameter budgets. In contrast, earlier methods either fine-tune the entire model (e.g., SHOT, AdaCon, DePT), requiring hundreds of millions of learnable parameters, or adapt a small subset (e.g., TENT, TPT) at the cost of retaining full computational graphs, often leading to weaker adaptation performance. These results highlight COLA’s balance between parameter efficiency and adaptability, motivating further investigation into its memory and runtime characteristics under practical deployment settings.

We evaluate COLA's computational efficiency during test-time adaptation on VisDA-C using ResNet-101 and ViT-B/16 backbones. Key metrics include activation memory, inference latency, and throughput. As shown in Table~\ref{tab:tta_efficiency}, COLA updates only 0.3\% of model parameters while improving accuracy from 85.0\% to 94.5\% with ResNet-101 and from 88.9\% to 98.3\% with ViT-B/16. During adaptation with batch size 128, it consumes only 0.7~GB of activation memory, which is 4\% of AdaCon’s 18.8~GB (batch size 64). The peak memory consumption during adaptation is consistently below 3.3 GB for both backbones. 
During inference, COLA adds less than 0.04~ms latency per image and retains over 95\% of CLIP's throughput. In contrast, prior parameter-efficient methods such as TENT and TPT require storing full computational graphs to update either distributed affine parameters or per-sample prompts, resulting in higher memory and runtime costs. For instance, COLA’s activation memory is less than 5\% of that required by TENT, and its inference is 57 times faster than TPT. These results confirm that COLA achieves superior adaptation performance with significantly lower computational cost.

\begin{table*}[!t]
  \centering
  \setlength{\tabcolsep}{3pt}
  \caption{Comparison of computational cost and performance of TTA methods on VisDA-C using ResNet-50.}
  \label{tab:tta_efficiency_rn50}
  \small
  \resizebox{2.\columnwidth}{!}{
  \begin{tabular}{l|c|c|c|c|c|c|c}
    \toprule
    Model & Backbone & Params (Tot / Trainable) & FLOPs (G) & Act. Mem (GB) & Infer. Time (ms) & Throughput (img/s) & Accuracy (\%) \\
    \midrule
    CLIP    & RN50  & 102.01M~/~-~~~ ~  & 57.34 & - & 0.79 & 1258.75 & 83.80 \\
    TENT    & RN50  & 25.56M~/~53.12K   & 24.79 & 10.28 & 1.72 & 581.22 & 70.70 \\
    SHOT  & RN50  & 24.04M~/~24.03M   & 8.26 & 10.28 & 0.67  & 1498.35   & 75.50 \\
    COLA    & RN50  & 102.64M~/~0.54M & 57.34 & 0.71 & 0.80 & 1252.45 & 96.70 \\
    \bottomrule
  \end{tabular}
  }
\end{table*}

\begin{table}[!t]
    \centering
    \setlength{\tabcolsep}{2pt}
    \small
    \caption{Classification accuracies (\%) for TTA on Terra Incognita. ``Loc.\,XX'' denotes diverse camera-trap locations.}
    \begin{tabular}{l|c|cccc|c}
    \toprule
          Method & Backbone &  Loc.38&  Loc.43&  Loc.46&  Loc.100& Avg.\\
          \midrule
          CLIP~\cite{radford2021learning}&RN-50&  25.9&  34.3&  24.2&  9.5& 23.5\\
          \rowcolor{Gray}
          COLA~(Ours)&RN-50&  \textbf{46.0}&  \textbf{44.6}&  29.1&  \textbf{18.6}& \textbf{34.6}\\
          \midrule
          CLIP~\cite{radford2021learning}&RN-101&  46.1&  39.6&  31.8&  \textbf{24.9}& 35.6\\
          \rowcolor{Gray}
          COLA~(Ours)&RN-101&  \textbf{46.4}&  \textbf{43.4}&  \textbf{33.5}&  23.4& \textbf{37.4}\\
          \midrule
          CLIP~\cite{radford2021learning}&ViT-B/16&  33.1&  37.5&  34.2&  \textbf{54.2}& 39.8\\
          \rowcolor{Gray}
          COLA~(Ours)&ViT-B/16&  \textbf{46.3}&  \textbf{45.5}&  \textbf{42.2}&  59.0& \textbf{48.3}\\
          \bottomrule
    \end{tabular}
    \label{tab:TerraIncognita}
\end{table}

To examine whether COLA’s efficiency extends to smaller backbones, we further evaluate its performance on CLIP with ResNet-50. As shown in Table~IX, COLA adds less than 0.01~ms of inference latency per image while improving VisDA-C accuracy by 12.9\% compared to the vanilla CLIP baseline. The activation memory required is only 0.71~GB, which is approximately 7\% of that used by TENT and SHOT at the same batch size. In terms of throughput, COLA achieves 2.13 times that of TENT and reaches 84\% of SHOT's throughput.
Notably, COLA achieves the highest accuracy among all compared methods. 
The reported resource usage for both CLIP and COLA includes a one-time invocation of the text encoder to compute class prototypes. This overhead is not incurred for each image during inference.
For reference, the CLIP image encoder (ResNet-50) contains 38.2 million parameters and requires 10.84~G FLOPs, while COLA introduces only 0.92 million additional FLOPs per image. These results suggest that COLA remains effective and resource-efficient even with limited-capacity encoders, reinforcing its potential for broader applicability.

\subsubsection{\textbf{Low Zero-shot Performance Scenarios}}
\paragraph{Setting} 
To assess COLA’s effectiveness under poor zero-shot performance conditions, we conduct additional experiments on the Terra Incognita~\cite{beery2018TerraIncognita}, which is a challenging wildlife recognition benchmark comprising approximately 11K images across 10 animal categories. 
The dataset includes four subsets (Locations 38, 43, 46, and 100), each collected by camera traps deployed at geographically distinct sites. 
These subsets differ markedly in lighting conditions, background context, and class composition. 
Owing to ecological factors, such as species-specific habitat preferences, the label distribution across locations is highly imbalanced. 
For example, Location 100 contains almost no samples from classes 0 and 1, further increasing the difficulty of domain adaptation.
\paragraph{Results}
As shown in Table~\ref{tab:TerraIncognita}, the initial zero-shot accuracies of CLIP on this dataset are relatively low: 23.5\% (RN-50), 35.6\% (RN-101), and 39.8\% (ViT-B/16). Despite these weak baselines, COLA consistently improves performance, achieving +11.1\%, +1.8\%, and +8.5\% gains respectively across the three backbones. The results show that COLA improves performance even on Terra Incognita, a scenario where CLIP’s initial zero-shot accuracy is relatively low, further highlighting the robustness of our method.

\begin{figure*}[!t]
  \centering
  \includegraphics[width=1\textwidth]
  {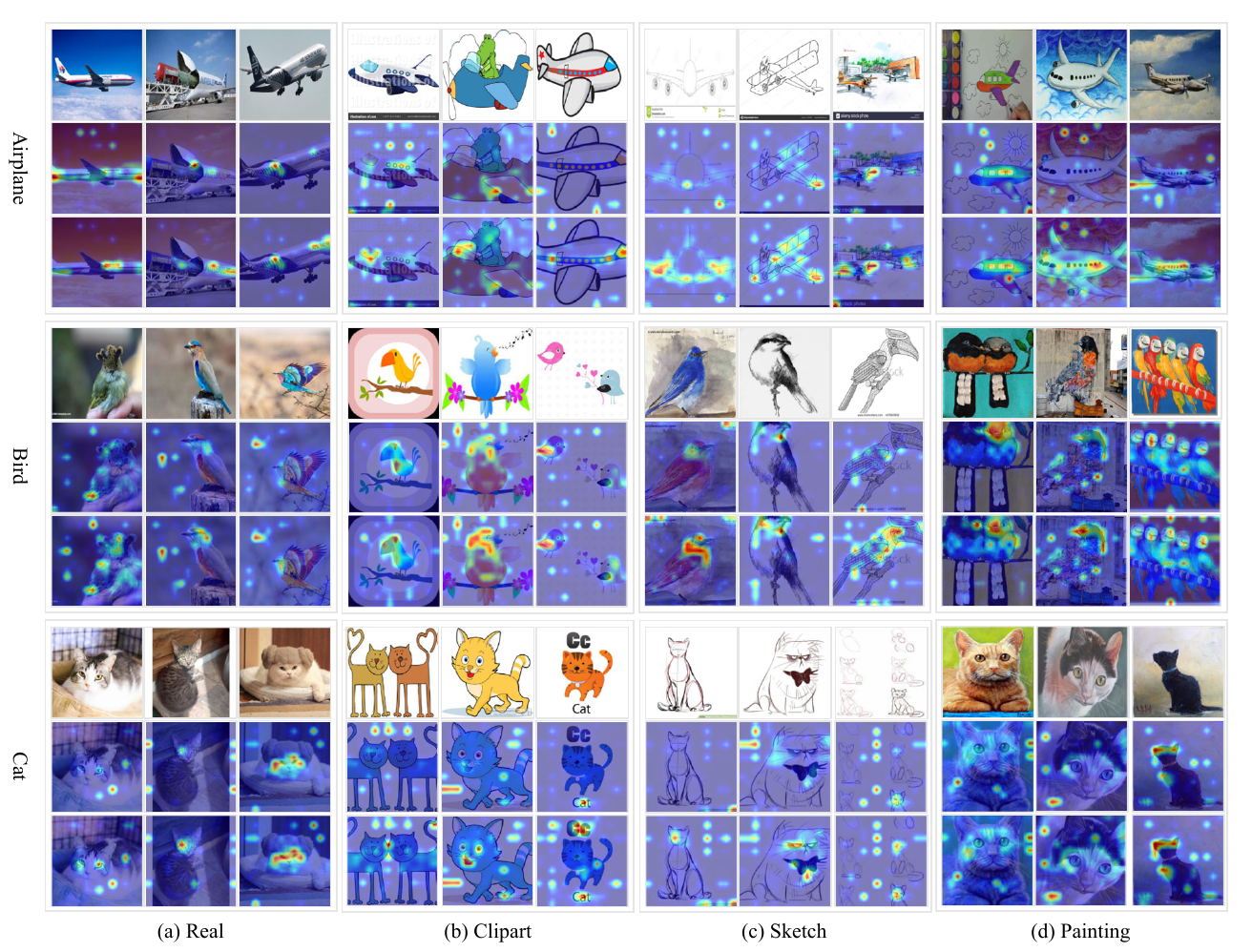}%
  \caption{Attention maps generated by CLIP and COLA features for ``airplane'', ``bird'', and ``cat'' across four domains in DomainNet-126. The first row for each category shows the original image, the second row displays the CLIP attention map, and the third row presents the COLA attention map. It is evident that the attention maps highlight different features for each category across the four domains. Additionally, for each category, COLA captures more relevant information and focuses on slightly different aspects within each domain.}

  \label{fig:Attention_Map}
\end{figure*}

\begin{table}[!t]
    \centering
    \setlength{\tabcolsep}{2pt}
    \caption{Ablation study of COLA for TTA task with ViT-B/16 Backbone. \#0 represents CLIP's zero-shot result.}
    \resizebox{1\columnwidth}{!}{
    \begin{tabular}{lcccccccc}
    \toprule
\multirow{2}{*}{\#} &\multirow{2}{*}{CLIP} & \multirow{2}{*}{Adapter} & \multicolumn{2}{c}{CAU} & \multirow{2}{*}{CBPL} & \multirow{2}{*}{VISDA-C} & \multirow{2}{*}{DomainNet-126} & \multirow{2}{*}{PACS}\\
& & &MLP & Mean & &\\ \midrule
        0 & \checkmark & ~ & ~ & ~ & ~ & 88.9 & 84.8 & 96.1\\ 
        1 & \checkmark & \checkmark & ~ & ~ &~ & 89.3 & 85.3 & 96.6 \\ 
        2 & \checkmark & \checkmark & \checkmark &~ &~ & 89.6 & 86.7 & 96.7 \\ 
        3 & \checkmark & & \checkmark & \checkmark & ~ & 96.8 & 85.7 & 97.1\\ 
        4 & \checkmark & \checkmark & \checkmark & \checkmark &~ & 97.6 & 86.8 & 97.6\\ 
        5 & \checkmark & \checkmark & \checkmark & \checkmark & \checkmark & \textbf{98.3} & \textbf{87.0} & \textbf{98.0}\\ 
        \bottomrule
    \end{tabular}
    }
\label{tab:ablation}
\end{table}

\subsubsection{\textbf{Ablation Study}}
\begin{table}[!t]
\centering
\caption{Coefficients for feature fusion on VisDA-C: $\alpha$ for CLIP features, $\beta$ for Adapter features, and $\gamma$ for CAU features.}
\label{tab:feature_fusion}
\resizebox{.95\columnwidth}{!}{
\begin{tabular}{cc|ccc|ccc|ccc}
\toprule
\multirow{2}{*}{$\gamma$} & \multicolumn{1}{c|}{$\alpha$} & \multicolumn{3}{c|}{0.1} & \multicolumn{3}{c|}{0.5} & \multicolumn{3}{c}{1} \\
    & $\beta$ & 0.1 & 0.5 & 1 & 0.1 & 0.5 & 1 & 0.1 & 0.5 & 1 \\
\midrule
0.1 & & 96.7 & 96.1 & 95.4 & 91.3 & 91.9 & 91.9 & 90.7 & 90.2 & 90.9 \\
0.5 & & 96.9 & 98.1 & 97.8 & 96.5 & 96.7 & 96.5 & 94.3 & 93.6 & 94.3 \\
1   & & 94.7 & \textbf{98.3} & 98.2 & 97.6 & 98.0 & 97.9 & 96.7 & 96.3 & 96.7 \\
\bottomrule
\end{tabular}
}
\end{table}
To analyze and validate the contributions of different components within COLA, we conducted ablation studies on VisDA-C, DomainNet-126, and PACS. As shown in Table~\ref{tab:ablation}, we begin with the CLIP zero-shot results, designated as baseline \#0, which achieve an average per-class accuracy of 88.9\% on VisDA-C, 84.8\% on DomainNet-126, and 96.1\% on PACS.
By appending an adapter after the image encoder of CLIP, we observe performance gains of 0.4\% on VisDA-C, 0.5\% on DomainNet-126, and 0.5\% on PACS. These improvements stem from the adapter’s ability to learn task-specific knowledge from the target domain.
To further examine the impact of domain-specific knowledge extraction, we decompose CAU into MLP and feature mean operations. In row \#2, we add an MLP in parallel, which achieves a 0.3\% gain on VisDA-C, 1.4\% on DomainNet-126, and 0.1\% on PACS. 
\begin{figure}[!t]
  \centering
\includegraphics[width=0.45\textwidth,]{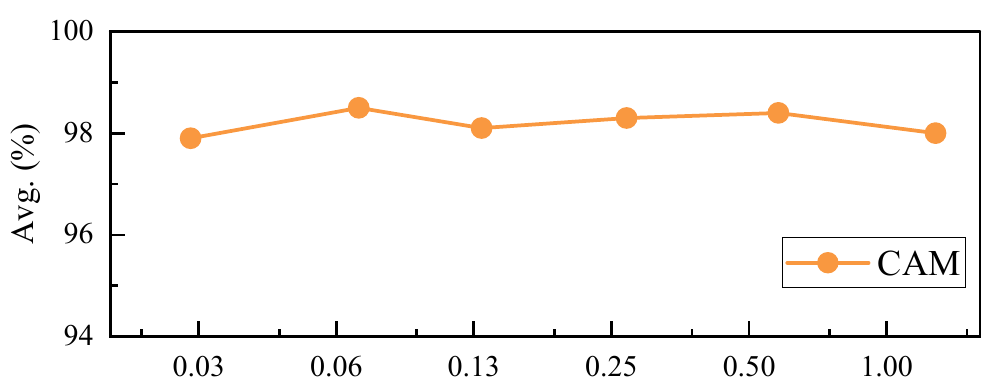}
  \caption{Effect of tunable parameter on COLA with ViT-B/16 for VisDA-C.}
  \label{fig:Robus_Para}
\end{figure}
In row \#3, we remove the adapter module and retain only the CAU, which consists of the mean operation and MLP. This prototype-only variant improves performance over the unadapted baseline, row \#0, by 7.9\% on VisDA-C, 0.9\% on DomainNet-126, and 1.0\% on PACS. However, it falls short of the full Context-Aware Module in row\#4 by 0.8\%, 1.1\%, and 0.5\%, respectively, indicating that the adapter provides essential task-specific knowledge guidance that complements the prototype mechanism.
We employ the full CAM structure in row \#4, which includes the adapter, MLP, and mean operation, increasing performance by 8.0\% to 97.6\% on VisDA-C, 0.1\% to 86.8\% on DomainNet-126, and 0.9\% to 97.6\% on PACS, attributed to the prototype mechanism capturing domain-wide patterns from batch-wise data.
In row \#5, we refine the pseudo-labeling scheme using the method introduced in Sec. III-B1. By balancing the quality and quantity of training pseudo labels, COLA further achieves 98.3\% on VisDA-C, 87.0\% on DomainNet-126, and 98.0\% on PACS.
Overall, the proposed components contribute differently across datasets, yet each consistently leads to performance gains when applied individually.

\subsubsection{\textbf{Feature Fusion}}

We analyzed the impact of different feature coefficients in Equation~\ref{eq:CLIP-CAM}. As shown in Table~\ref{tab:feature_fusion}, when alpha and beta are fixed, accuracy increases with gamma, indicating that the features from CAU significantly aid the model in adapting to the target domain. 
When beta and gamma are fixed, an increase in alpha leads to a decrease in performance; however, the lowest observed performance still surpasses that of CLIP's zero-shot. This suggests that a higher proportion of CLIP features gradually limits the model’s adaptability.
When alpha and gamma are fixed, the optimal performance typically occurs at a beta value of 0.5, indicating that the features learned by the Adapter help the model adapt to the new domain, but with limitations.

Additionally, Table~\ref{tab:feature_fusion} also shows that the optimal coefficient combination for performance is alpha=0.1, beta=0.5, and gamma=1, whereas the weakest combination is alpha=1, beta=0.5, and gamma=0.1. This further validates the critical role of CAU features in the feature fusion process, while an excessive weight of CLIP features may limit the model’s adaptability.

\begin{table}[!t]
    \centering
    \caption{{Average accuracy (\%) of COLA with different pseudo-labeling strategies on Terra Incognita.}}
    \label{tab:TerraIncognita_pl}
    \begin{tabular}{l|cccc}
    \toprule
    Backbone & CLIP & Basic & GPUE & \cellcolor{gray!20} \textbf{COLA (Ours)} \\
    \midrule
    RN-50    & 23.5 & 22.9 & 28.0 & \cellcolor{gray!20} \textbf{34.6} \\
    RN-101   & 35.6 & 35.5 & 34.6 & \cellcolor{gray!20} \textbf{36.7} \\
    ViT-B/16 & 39.8 & 48.1 & 47.6 & \cellcolor{gray!20} \textbf{48.3} \\
    \bottomrule
    \end{tabular}
\end{table}

\subsubsection{\textbf{Pseudo-Labeling Strategies}}
To further assess COLA’s performance when pseudo-label reliability is severely limited, we conducted experiments on the Terra Incognita dataset  (Table~\ref{tab:TerraIncognita_pl}).
This benchmark features substantial domain shifts and a high level of label noise, resulting in low-quality supervision from the backbone model. CLIP’s zero-shot accuracy ranges from only 23.5\% to 39.8\%. These conditions present a particularly challenging testbed for adaptation algorithms.

We compared three pseudo-labeling strategies under consistent experimental settings: using a fixed confidence threshold 0.9, advanced GPUE approach~\cite{litrico2023_PLUE} based on uncertainty estimation, and our CBPL.
As shown in Table~\ref{tab:TerraIncognita_pl}, the results reveal the following key findings: 
(i) Basic confidence thresholding yielded inconsistent results, with accuracy changes of $-0.6\%$, $-0.1\%$, and $8.3\%$ across different backbones, highlighting its limited reliability under noisy conditions. 
(ii) The simplified GPUE-inspired refinement, which incorporates uncertainty-guided loss reweighting and memory-bank–based neighbour refinement, achieved moderate gains of $4.5\%$, $-1.0\%$, and $7.8\%$, but at the cost of high activation memory (peak 5334~MiB). 
(iii) Our CBPL strategy, consistently outperformed the baselines ($11.1\%$, $1.1\%$, $8.5\%$) while substantially reducing memory usage to 1566~MiB.
These results demonstrate that CBPL not only improves pseudo-label quality and adaptation robustness but also offers a favourable trade-off between performance and efficiency compared to alternative strategies.

\begin{figure}[!t]
  \centering
  \includegraphics[width=0.5\textwidth]{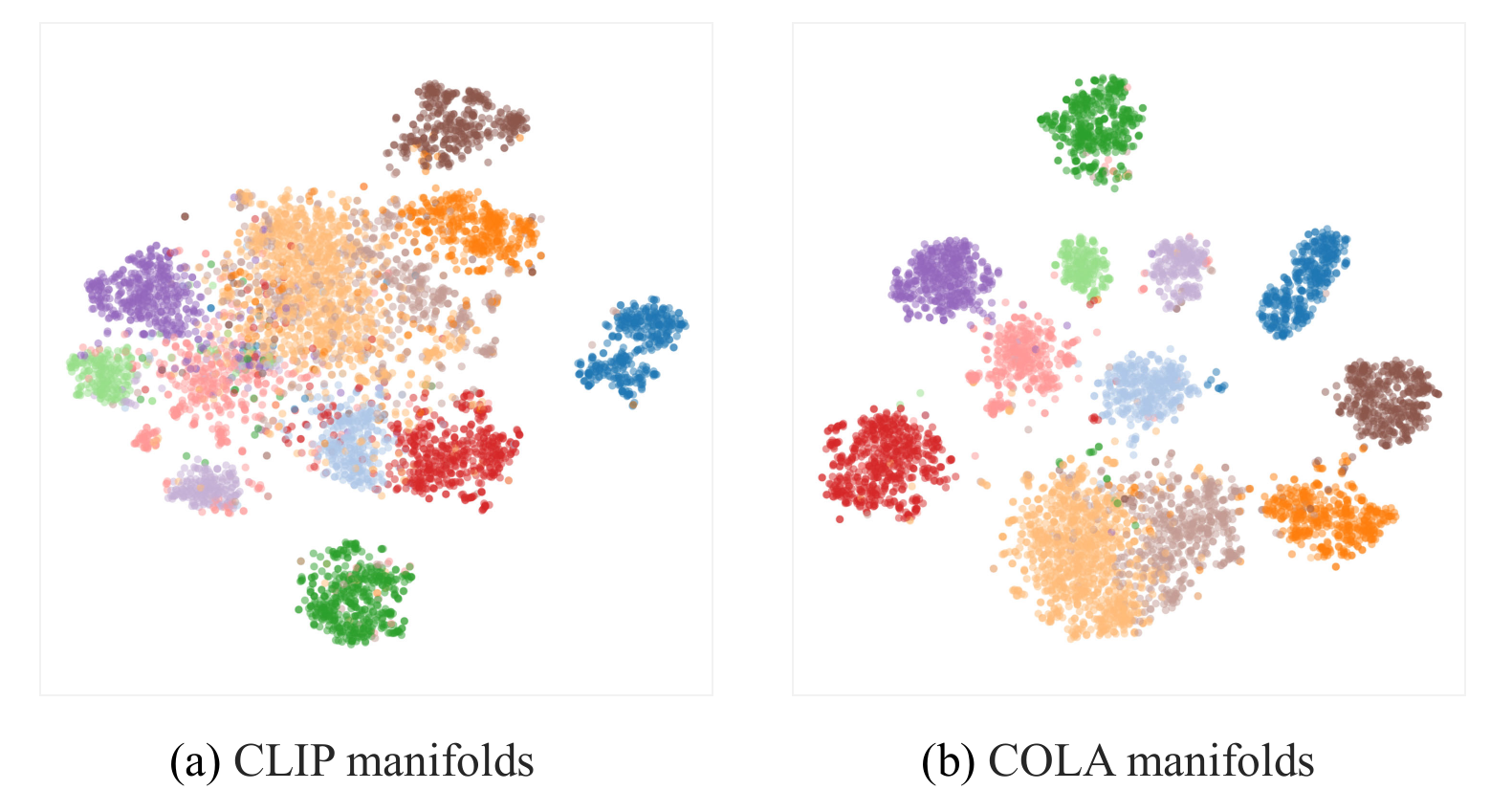}
  \caption{Visualisation of CLIP and COLA manifolds for TTA on ViSDA-C via t-SNE. It is clear that COLA shows a more obvious separation of image features compared with CLIP.}
  \label{fig:TSNE}
\end{figure}

\subsubsection{\textbf{Sensitivity}}

This section delineates the impacts of varying quantities of adjustable parameters on the overall performance and the effects when feature fusion is applied.
\label{sec:S_Sen}

Figure~\ref{fig:Robus_Para} illustrates the performance of the proposed COLA method under various parameter configurations, reflecting the method's sensitivity to changes in the number of trainable parameters. The number of neurons in the adapter and the hidden layers of the CAU was systematically increased from 16 to 512, methodically demonstrating the impact of these parameter adjustments on performance metrics. The findings affirm that the method's efficacy is consistently maintained across diverse configurations. Furthermore, altering the number of MLP layers in the CAU from two to four, while keeping the adapter settings fixed, was shown to minimally affect overall performance, with metrics at 98.3\%, 98.3\%, and 98.2\%, respectively.

\begin{figure}[!t]
  \centering
  \includegraphics[width=0.5\textwidth]{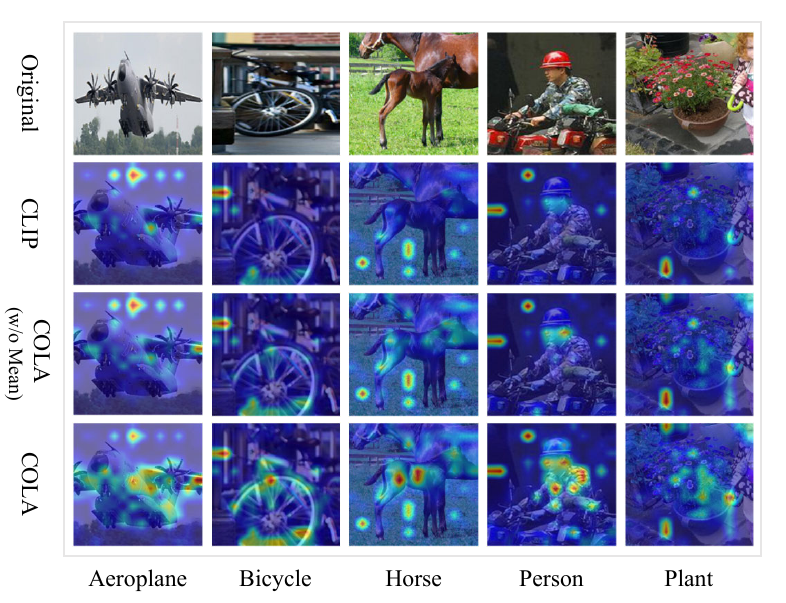}
  \caption{Visualisation of attention maps on VisDA-C Using ViT-B/16. We selected several representative categories from the VisDA-C dataset. We compared the attention maps generated by CLIP, COLA without the ``mean computation'' module, and full COLA. Note that ``w/o'' denotes ``without''.}
  \label{fig:COAL_VISDA}
\end{figure}
 
 \subsubsection{\textbf{Visualization}}

We compare the attention maps from CLIP, COLA without the mean calculation module, and full COLA across the categories ``Aeroplane'', ``Bicycle'', ``Horse'', ``Person'', and ``Plant'' in VisDA-C, as shown in Figure~\ref{fig:COAL_VISDA}. Compared to CLIP's attention maps, COLA without the mean calculation module slightly increases attention towards the target category. In contrast, full COLA significantly improves the model's focus on the target category. This finding suggests that processing mean features makes it easier to identify key features of the target category. Moreover, these processed features help the model adapt better to the target domain.

Subsequently, we compare the attention maps from CLIP and COLA features for the categories ``airplane'', ``bird'', and ``cat'' across four domains in DomainNet-126, as shown in Figure~\ref{fig:Attention_Map}. It is evident that COLA learns both meaningful domain-specific and task-specific knowledge.
 Specifically, COLA mines and utilizes latent information relevant to the target domain, and better focuses on the information of the target categories. 

 For instance, when identifying the category ``aircraft'', COLA focuses on features such as text, logos, wings, and windows. Similarly, when identifying the category ``bird'', COLA concentrates on features such as the head, neck, beak, and feathers. In the case of identifying the category ``cat'', COLA emphasizes features such as the eyes, body, text, patterns, and other distinctive characteristics.
 Although the emphasis differs across domains, COLA consistently targets features pertinent to each category.

Further, we utilise t-SNE to visualise the CLIP and COLA manifolds on VisDA-C, shown in Figure~\ref{fig:TSNE}, where 12 colours denote 12 classes, respectively. It is clearly illustrated that in high-dimensional classification space, our method COLA, shows a much more obvious separation of image features belonging to different categories. For confusing categories such as car (light orange points) and truck (pale brown points), compared with CLIP, COLA is more effective in detecting the similarities among the image manifolds from the same class. In summary, the visualisation results prove that COLA excels at learning better feature representation under the TTA setup.

\section{Conclusion}
To overcome the limitation of existing TTAs that the source domain and the target domain share the same label space, we propose a unified framework called COLA, which can be easily adapted to any target domain. 
We find that a vision language model like CLIP suits naturally for TTA due to its great zero-shot performance and its flexible text-image alignment classifier. 
To effectively and efficiently use CLIP, a light-weight context-aware module is designed to learn novel information from the target domain. 
To make full use of knowledge acquired in CLIP, a CBPL scheme is adopted to filter accurate and diverse samples to train the context-aware module alone.
Experiments show that COLA significantly improves the performance on major TTA benchmarks.
Besides, we find that our COLA has a strong generalization ability compared with other CLIP-based models.

\section*{Acknowledgments}
This work was supported in part by the National Natural Science Foundation of China under Grant No.62206305.

\bibliographystyle{IEEEtran}
\bibliography{References}

\begin{thebibliography}{10}
\providecommand{\url}[1]{#1}
\csname url@samestyle\endcsname
\providecommand{\newblock}{\relax}
\providecommand{\bibinfo}[2]{#2}
\providecommand{\BIBentrySTDinterwordspacing}{\spaceskip=0pt\relax}
\providecommand{\BIBentryALTinterwordstretchfactor}{4}
\providecommand{\BIBentryALTinterwordspacing}{\spaceskip=\fontdimen2\font plus
\BIBentryALTinterwordstretchfactor\fontdimen3\font minus \fontdimen4\font\relax}
\providecommand{\BIBforeignlanguage}[2]{{%
\expandafter\ifx\csname l@#1\endcsname\relax
\typeout{** WARNING: IEEEtran.bst: No hyphenation pattern has been}%
\typeout{** loaded for the language `#1'. Using the pattern for}%
\typeout{** the default language instead.}%
\else
\language=\csname l@#1\endcsname
\fi
#2}}
\providecommand{\BIBdecl}{\relax}
\BIBdecl

\bibitem{wei2018person}
L.~Wei, S.~Zhang, W.~Gao, and Q.~Tian, ``Person transfer gan to bridge domain gap for person re-identification,'' in \emph{CVPR}, 2018, pp. 79--88.

\bibitem{cordts2016cityscapes}
M.~Cordts, M.~Omran, S.~Ramos, T.~Rehfeld, M.~Enzweiler, R.~Benenson, U.~Franke, S.~Roth, and B.~Schiele, ``The cityscapes dataset for semantic urban scene understanding,'' in \emph{CVPR}, 2016, pp. 3213--3223.

\bibitem{guan2021domain}
H.~Guan and M.~Liu, ``Domain adaptation for medical image analysis: a survey,'' \emph{IEEE TBME}, vol.~69, no.~3, pp. 1173--1185, 2021.

\bibitem{patel2015visual}
V.~M. Patel, R.~Gopalan, R.~Li, and R.~Chellappa, ``Visual domain adaptation: A survey of recent advances,'' \emph{SPM}, vol.~32, no.~3, pp. 53--69, 2015.

\bibitem{long2015learning}
M.~Long, Y.~Cao, J.~Wang, and M.~Jordan, ``Learning transferable features with deep adaptation networks,'' in \emph{ICML}, 2015, pp. 97--105.

\bibitem{lu2020stochastic}
Z.~Lu, Y.~Yang, X.~Zhu, C.~Liu, Y.-Z. Song, and T.~Xiang, ``Stochastic classifiers for unsupervised domain adaptation,'' in \emph{CVPR}, 2020, pp. 9111--9120.

\bibitem{liang2020we}
J.~Liang, D.~Hu, and J.~Feng, ``Do we really need to access the source data? source hypothesis transfer for unsupervised domain adaptation,'' in \emph{ICML}, 2020, pp. 6028--6039.

\bibitem{Chen_2022_CVPR}
D.~Chen, D.~Wang, T.~Darrell, and S.~Ebrahimi, ``Contrastive test-time adaptation,'' in \emph{CVPR}, 2022, pp. 295--305.

\bibitem{gao2022_DEPT}
Y.~Gao, X.~Shi, Y.~Zhu, H.~Wang, Z.~Tang, X.~Zhou, M.~Li, and D.~N. Metaxas, ``Visual prompt tuning for test-time domain adaptation,'' \emph{arXiv preprint arXiv:2210.04831}, 2022.

\bibitem{lu2023uncertainty}
Z.~Lu, D.~Li, Y.-Z. Song, T.~Xiang, and T.~M. Hospedales, ``Uncertainty-aware source-free domain adaptive semantic segmentation,'' \emph{TIP}, 2023.

\bibitem{radford2021learning}
A.~Radford, J.~W. Kim, C.~Hallacy, A.~Ramesh, G.~Goh, S.~Agarwal, G.~Sastry, A.~Askell, P.~Mishkin, J.~Clark \emph{et~al.}, ``Learning transferable visual models from natural language supervision,'' in \emph{ICML}, 2021, pp. 8748--8763.

\bibitem{liang2023comprehensive}
J.~Liang, R.~He, and T.~Tan, ``A comprehensive survey on test-time adaptation under distribution shifts,'' \emph{arXiv preprint arXiv:2303.15361}, 2023.

\bibitem{yang2021_NRC}
S.~Yang, J.~van~de Weijer, L.~Herranz, S.~Jui \emph{et~al.}, ``Exploiting the intrinsic neighborhood structure for source-free domain adaptation,'' \emph{NeurIPS}, vol.~34, pp. 29\,393--29\,405, 2021.

\bibitem{wang2022_CDCL}
R.~Wang, Z.~Wu, Z.~Weng, J.~Chen, G.-J. Qi, and Y.-G. Jiang, ``Cross-domain contrastive learning for unsupervised domain adaptation,'' \emph{TMM}, 2022.

\bibitem{wang2020tent}
D.~Wang, E.~Shelhamer, S.~Liu, B.~Olshausen, and T.~Darrell, ``Tent: Fully test-time adaptation by entropy minimization,'' in \emph{ICLR}, 2020.

\bibitem{tian2021vdm}
J.~Tian, J.~Zhang, W.~Li, and D.~Xu, ``Vdm-da: Virtual domain modeling for source data-free domain adaptation,'' \emph{IEEE TCSVT}, vol.~32, no.~6, pp. 3749--3760, 2021.

\bibitem{qiu2021_CPGA}
Z.~Qiu, Y.~Zhang, H.~Lin, S.~Niu, Y.~Liu, Q.~Du, and M.~Tan, ``Source-free domain adaptation via avatar prototype generation and adaptation,'' \emph{IJCAI}, 2021.

\bibitem{liang2021shotplus}
J.~Liang, D.~Hu, Y.~Wang, R.~He, and J.~Feng, ``Source data-absent unsupervised domain adaptation through hypothesis transfer and labeling transfer,'' \emph{IEEE TPAMI}, vol.~44, no.~11, pp. 8602--8617, 2021.

\bibitem{ahmed2022cross}
S.~M. Ahmed, S.~Lohit, K.-C. Peng, M.~J. Jones, and A.~K. Roy-Chowdhury, ``Cross-modal knowledge transfer without task-relevant source data,'' in \emph{ECCV}, 2022, pp. 111--127.

\bibitem{lao2021hypothesis}
Q.~Lao, X.~Jiang, and M.~Havaei, ``Hypothesis disparity regularized mutual information maximization,'' in \emph{AAAI}, vol.~35, no.~9, 2021, pp. 8243--8251.

\bibitem{wang2022exploring}
F.~Wang, Z.~Han, Y.~Gong, and Y.~Yin, ``Exploring domain-invariant parameters for source free domain adaptation,'' in \emph{ECCV}, 2022, pp. 7151--7160.

\bibitem{ramesh2022Dalle2}
A.~Ramesh, P.~Dhariwal, A.~Nichol, C.~Chu, and M.~Chen, ``Hierarchical text-conditional image generation with clip latents,'' \emph{arXiv preprint arXiv:2204.06125}, 2022.

\bibitem{luo2022clip4clip}
H.~Luo, L.~Ji, M.~Zhong, Y.~Chen, W.~Lei, N.~Duan, and T.~Li, ``Clip4clip: An empirical study of clip for end to end video clip retrieval and captioning,'' \emph{Neurocomputing}, vol. 508, pp. 293--304, 2022.

\bibitem{guzhov2022audioclip}
A.~Guzhov, F.~Raue, J.~Hees, and A.~Dengel, ``Audioclip: Extending clip to image, text and audio,'' in \emph{ICASSP}, 2022, pp. 976--980.

\bibitem{Wang_2022_CVPR_CLIPseg}
Z.~Wang, Y.~Lu, Q.~Li, X.~Tao, Y.~Guo, M.~Gong, and T.~Liu, ``Cris: Clip-driven referring image segmentation,'' in \emph{CVPR}, 2022, pp. 11\,686--11\,695.

\bibitem{gu2021open}
X.~Gu, T.-Y. Lin, W.~Kuo, and Y.~Cui, ``Open-vocabulary object detection via vision and language knowledge distillation,'' in \emph{ICLR}, 2021.

\bibitem{zhang2023_VLP_vision}
J.~Zhang, J.~Huang, S.~Jin, and S.~Lu, ``Vision-language models for vision tasks: A survey,'' \emph{arXiv preprint arXiv:2304.00685}, 2023.

\bibitem{gan2022_VLP_survey}
Z.~Gan, L.~Li, C.~Li, L.~Wang, Z.~Liu, J.~Gao \emph{et~al.}, ``Vision-language pre-training: Basics, recent advances, and future trends,'' \emph{Found. Trends Comput. Graph. Vis.}, vol.~14, no. 3--4, pp. 163--352, 2022.

\bibitem{devlin-etal-2019-bert}
J.~Devlin, M.-W. Chang, K.~Lee, and K.~Toutanova, ``Bert: Pre-training of deep bidirectional transformers for language understanding,'' in \emph{NAACL}, 2019, pp. 4171--4186.

\bibitem{he2020_MoCo}
K.~He, H.~Fan, Y.~Wu, S.~Xie, and R.~Girshick, ``Momentum contrast for unsupervised visual representation learning,'' in \emph{CVPR}, 2020, pp. 9729--9738.

\bibitem{SimCLR-chen20j}
T.~Chen, S.~Kornblith, M.~Norouzi, and G.~Hinton, ``A simple framework for contrastive learning of visual representations,'' in \emph{ICML}, vol. 119, 2020, pp. 1597--1607.

\bibitem{jia2021_ALIGN}
C.~Jia, Y.~Yang, Y.~Xia, Y.-T. Chen, Z.~Parekh, H.~Pham, Q.~Le, Y.-H. Sung, Z.~Li, and T.~Duerig, ``Scaling up visual and vision-language representation learning with noisy text supervision,'' in \emph{ICML}, 2021, pp. 4904--4916.

\bibitem{schuhmann2022laion}
C.~Schuhmann, R.~Beaumont, R.~Vencu, C.~Gordon, R.~Wightman, M.~Cherti, T.~Coombes, A.~Katta, C.~Mullis, M.~Wortsman \emph{et~al.}, ``Laion-5b: An open large-scale dataset for training next generation image-text models,'' \emph{NeurIPS}, vol.~35, pp. 25\,278--25\,294, 2022.

\bibitem{gu2022wukong}
J.~Gu, X.~Meng, G.~Lu, L.~Hou, N.~Minzhe, X.~Liang, L.~Yao, R.~Huang, W.~Zhang, X.~Jiang \emph{et~al.}, ``Wukong: A 100 million large-scale chinese cross-modal pre-training benchmark,'' \emph{NeurIPS}, vol.~35, pp. 26\,418--26\,431, 2022.

\bibitem{liu2024few}
F.~Liu, T.~Zhang, W.~Dai, C.~Zhang, W.~Cai, X.~Zhou, and D.~Chen, ``Few-shot adaptation of multi-modal foundation models: A survey,'' \emph{Artificial Intelligence Review}, vol.~57, no.~10, p. 268, 2024.

\bibitem{ding2023parameter}
N.~Ding, Y.~Qin, G.~Yang, F.~Wei, Z.~Yang, Y.~Su, S.~Hu, Y.~Chen, C.-M. Chan, W.~Chen \emph{et~al.}, ``Parameter-efficient fine-tuning of large-scale pre-trained language models,'' \emph{Nature Machine Intelligence}, vol.~5, no.~3, pp. 220--235, 2023.

\bibitem{lialin2023scaling}
V.~Lialin, V.~Deshpande, and A.~Rumshisky, ``Scaling down to scale up: A guide to parameter-efficient fine-tuning,'' \emph{arXiv preprint arXiv:2303.15647}, 2023.

\bibitem{ding2022delta}
N.~Ding, Y.~Qin, G.~Yang, F.~Wei, Z.~Yang, Y.~Su, S.~Hu, Y.~Chen, C.-M. Chan, W.~Chen \emph{et~al.}, ``Delta tuning: A comprehensive study of parameter efficient methods for pre-trained language models,'' \emph{arXiv preprint arXiv:2203.06904}, 2022.

\bibitem{zhou2022CoOp}
K.~Zhou, J.~Yang, C.~C. Loy, and Z.~Liu, ``Learning to prompt for vision-language models,'' \emph{IJCV}, vol. 130, no.~9, pp. 2337--2348, 2022.

\bibitem{zhou2022_CoCoOp}
K.~Zhou, J.~Yang, C.~C. Loy, and Z.~Liu, ``Conditional prompt learning for vision-language models,'' in \emph{CVPR}, 2022, pp. 16\,816--16\,825.

\bibitem{zhu2023_ProGrad}
B.~Zhu, Y.~Niu, Y.~Han, Y.~Wu, and H.~Zhang, ``Prompt-aligned gradient for prompt tuning,'' in \emph{CVPR}, 2023, pp. 15\,659--15\,669.

\bibitem{ben-zaken-etal-2022-bitfit}
E.~Ben~Zaken, Y.~Goldberg, and S.~Ravfogel, ``{B}it{F}it: Simple parameter-efficient fine-tuning for transformer-based masked language-models,'' in \emph{ACL}, vol.~2, 2022, pp. 1--9.

\bibitem{gao2021clip}
P.~Gao, S.~Geng, R.~Zhang, T.~Ma, R.~Fang, Y.~Zhang, H.~Li, and Y.~Qiao, ``Clip-adapter: Better vision-language models with feature adapters,'' \emph{IJCV}, pp. 1--15, 2021.

\bibitem{yu2023task}
T.~Yu, Z.~Lu, X.~Jin, Z.~Chen, and X.~Wang, ``Task residual for tuning vision-language models,'' in \emph{CVPR}, 2023, pp. 10\,899--10\,909.

\bibitem{li2024graphadapter}
X.~Li, D.~Lian, Z.~Lu, J.~Bai, Z.~Chen, and X.~Wang, ``Graphadapter: Tuning vision-language models with dual knowledge graph,'' \emph{NeurIPS}, vol.~36, 2023.

\bibitem{lu2023beyond}
Z.~Lu, J.~Bai, X.~Li, Z.~Xiao, and X.~Wang, ``Beyond sole strength: Customized ensembles for generalized vision-language models,'' in \emph{ICML}, 2024.

\bibitem{houlsby2019parameter}
N.~Houlsby, A.~Giurgiu, S.~Jastrzebski, B.~Morrone, Q.~De~Laroussilhe, A.~Gesmundo, M.~Attariyan, and S.~Gelly, ``Parameter-efficient transfer learning for nlp,'' in \emph{ICML}, 2019, pp. 2790--2799.

\bibitem{shu2022TPT}
M.~Shu, W.~Nie, D.-A. Huang, Z.~Yu, T.~Goldstein, A.~Anandkumar, and C.~Xiao, ``Test-time prompt tuning for zero-shot generalization in vision-language models,'' \emph{NeurIPS}, vol.~35, pp. 14\,274--14\,289, 2022.

\bibitem{chen2022_SSNLL}
W.~Chen, L.~Lin, S.~Yang, D.~Xie, S.~Pu, and Y.~Zhuang, ``Self-supervised noisy label learning for source-free unsupervised domain adaptation,'' in \emph{IROS}, 2022, pp. 10\,185--10\,192.

\bibitem{saito2019semi}
K.~Saito, D.~Kim, S.~Sclaroff, T.~Darrell, and K.~Saenko, ``Semi-supervised domain adaptation via minimax entropy,'' in \emph{ICCV}, 2019, pp. 8050--8058.

\bibitem{karim2023_C-SFDA}
N.~Karim, N.~C. Mithun, A.~Rajvanshi, H.-p. Chiu, S.~Samarasekera, and N.~Rahnavard, ``C-sfda: A curriculum learning aided self-training framework for efficient source free domain adaptation,'' in \emph{CVPR}, 2023, pp. 24\,120--24\,131.

\bibitem{peng2017visda}
X.~Peng, B.~Usman, N.~Kaushik, J.~Hoffman, D.~Wang, and K.~Saenko, ``Visda: The visual domain adaptation challenge,'' \emph{arXiv preprint arXiv:1710.06924}, 2017.

\bibitem{khosla2012undoing_VLCS}
A.~Khosla, T.~Zhou, T.~Malisiewicz, A.~A. Efros, and A.~Torralba, ``Undoing the damage of dataset bias,'' in \emph{ECCV}.\hskip 1em plus 0.5em minus 0.4em\relax Springer, 2012, pp. 158--171.

\bibitem{li2017deeper_PACS}
D.~Li, Y.~Yang, Y.-Z. Song, and T.~M. Hospedales, ``Deeper, broader and artier domain generalization,'' in \emph{CVPR}, 2017, pp. 5542--5550.

\bibitem{venkateswara2017_OfficeHome}
H.~Venkateswara, J.~Eusebio, S.~Chakraborty, and S.~Panchanathan, ``Deep hashing network for unsupervised domain adaptation,'' in \emph{CVPR}, 2017, pp. 5018--5027.

\bibitem{xian2017zero_harmonic}
Y.~Xian, B.~Schiele, and Z.~Akata, ``Zero-shot learning-the good, the bad and the ugly,'' in \emph{CVPR}, 2017, pp. 4582--4591.

\bibitem{fei2004learning_Caltech101}
L.~Fei-Fei, R.~Fergus, and P.~Perona, ``Learning generative visual models from few training examples: An incremental bayesian approach tested on 101 object categories,'' in \emph{CVPRW}.\hskip 1em plus 0.5em minus 0.4em\relax IEEE, 2004, pp. 178--178.

\bibitem{parkhi2012cats_Pets}
O.~M. Parkhi, A.~Vedaldi, A.~Zisserman, and C.~Jawahar, ``Cats and dogs,'' in \emph{CVPR}.\hskip 1em plus 0.5em minus 0.4em\relax IEEE, 2012, pp. 3498--3505.

\bibitem{nilsback2008automated_Flowers102}
M.-E. Nilsback and A.~Zisserman, ``Automated flower classification over a large number of classes,'' in \emph{ICVGIP}.\hskip 1em plus 0.5em minus 0.4em\relax IEEE, 2008, pp. 722--729.

\bibitem{taori2020measuring_UCF101}
R.~Taori, A.~Dave, V.~Shankar, N.~Carlini, B.~Recht, and L.~Schmidt, ``Measuring robustness to natural distribution shifts in image classification,'' \emph{NeurIPS}, vol.~33, pp. 18\,583--18\,599, 2020.

\bibitem{zhu2022thop}
L.~Zhu, ``Thop: Pytorch-opcounter,'' \emph{THOP: PyTorch-OpCounter}, 2022.

\bibitem{beery2018TerraIncognita}
S.~Beery, G.~Van~Horn, and P.~Perona, ``Recognition in terra incognita,'' in \emph{ECCV}, 2018, pp. 456--473.

\bibitem{litrico2023_PLUE}
M.~Litrico, A.~Del~Bue, and P.~Morerio, ``Guiding pseudo-labels with uncertainty estimation for source-free unsupervised domain adaptation,'' in \emph{CVPR}, 2023, pp. 7640--7650.

\end{thebibliography}

\end{document}